\documentclass[letterpaper, 10 pt, conference]{ieeeconf}

\IEEEoverridecommandlockouts                              %
\usepackage{amsmath,amsfonts,bm}

\def\eqref#1{equation~\ref{#1}}

\def\1{\bm{1}}

\DeclareMathAlphabet{\mathsfit}{\encodingdefault}{\sfdefault}{m}{sl}
\SetMathAlphabet{\mathsfit}{bold}{\encodingdefault}{\sfdefault}{bx}{n}

\newcommand{\E}{\mathbb{E}}

\newcommand{\new}[1]{\textcolor{black}{#1}}
\newcommand{\corlnew}[1]{\textcolor{black}{#1}}
\newcommand{\method}{SPRINT}

\newcommand{\ie}{i.e.,\ }
\newcommand{\eg}{e.g.,\ }
\newcommand{\myfig}[1]{Figure~\ref{#1}}

\newcommand{\mytable}[1]{Table~\ref{#1}}
\newcommand{\myeq}[1]{Eq.~\ref{#1}}
\newcommand{\mysec}[1]{Section~\ref{#1}}

\usepackage[backend=biber,style=ieee,natbib=true]{biblatex} 
\renewcommand{\bibfont}{\footnotesize} %
\usepackage{hyperref}
\usepackage{graphicx}
\usepackage{dsfont}

\usepackage{float}
\usepackage{wrapfig}
\usepackage{url}
\usepackage{bbm} %
\usepackage{booktabs} %
\usepackage{subcaption} %
\usepackage{mdframed}
\usepackage{algorithm, algorithmicx}
\usepackage[noend]{algpseudocode}
\usepackage{multirow}
\usepackage{anyfontsize}
\hypersetup{
breaklinks=true,letterpaper=true,colorlinks,citecolor=blue
    }

\definecolor{sprint}{RGB}{50,150,77}

\title{\method: Scalable Policy Pre-Training via Language Instruction Relabeling}
\author{Jesse Zhang$^{1}$, Karl Pertsch$^{1}$, Jiahui Zhang$^{1}$, Joseph J. Lim$^{2}$%
\thanks{$^{1}$University of Southern California, $^{2}$Korea Advanced Institute of Science and Technology}
}
\begin{document}
\makeatletter
\let\@oldmaketitle\@maketitle%
\renewcommand{\@maketitle}{\@oldmaketitle%
    \includegraphics[width=\linewidth]{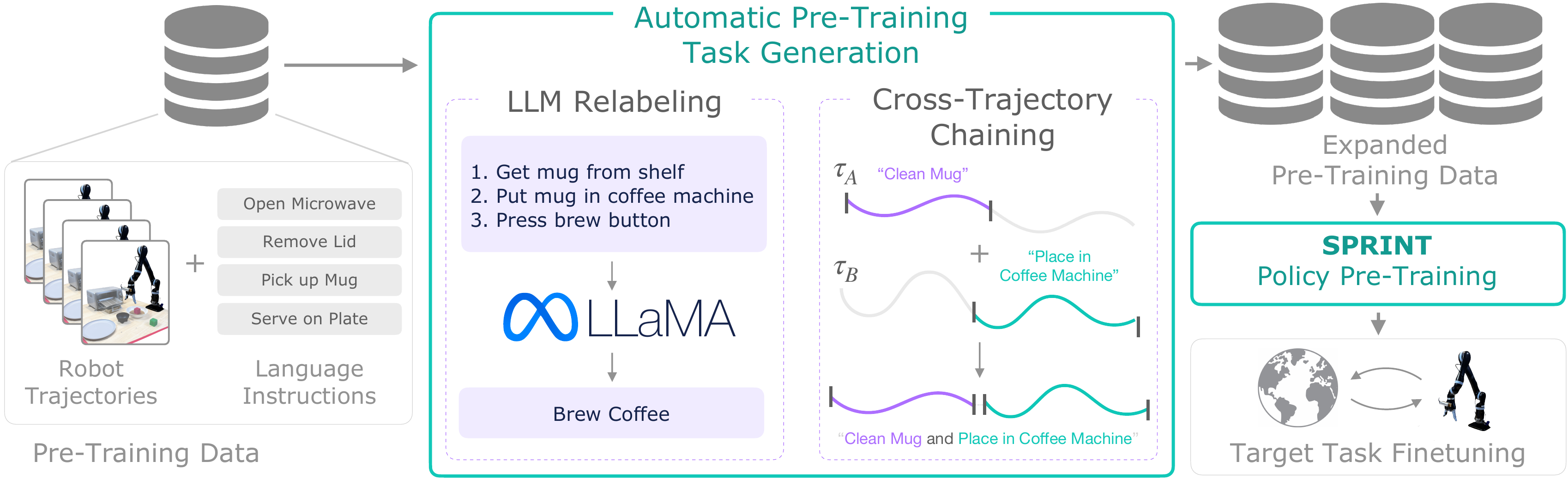} \\[0.25em]
   \refstepcounter{figure}\footnotesize{{Fig. 1:} \method\ is a scalable approach for pre-training robot policies with a rich repertoire of skills while minimizing human annotation effort. Given a dataset of language-annotated trajectories for offline pre-training, \method\ automatically expands the skill set via LLM-based \textbf{instruction relabeling} and \textbf{cross-trajectory skill chaining} to enable efficient finetuning on unseen target tasks.}
  \label{fig:teaser} \medskip \vspace{-10pt}}%
\makeatother

\maketitle
\renewcommand\thefigure{\arabic{figure}}
\setcounter{figure}{1}

\begin{abstract}
Pre-training robots with a rich set of skills can substantially accelerate the learning of downstream tasks. Prior works have defined pre-training tasks via natural language instructions, but doing so requires tedious human annotation of hundreds of thousands of instructions. Thus, we propose SPRINT, a scalable offline policy pre-training approach which substantially reduces the human effort needed for pre-training a diverse set of skills. Our method uses two core ideas to automatically expand a base set of pre-training tasks: instruction relabeling via large language models and cross-trajectory \textit{skill chaining} with offline reinforcement learning. As a result, SPRINT pre-training equips robots with a richer repertoire of skills \corlnew{that can help an agent generalize to new tasks.} Experiments in a household simulator and on a real robot kitchen manipulation task show that SPRINT leads to substantially faster learning of new long-horizon tasks than previous pre-training approaches. %
Website at \href{https://clvrai.com/sprint}{https://clvrai.com/sprint}.
\end{abstract}

\section{Introduction}

When humans learn a new task, \eg how to cook a new dish, we rely on a large repertoire of previously learned \textit{skills}, like ``\textit{chopping vegetables}" or ``\textit{boiling pasta}", that make learning more efficient.
Similarly, much work in robot learning aims to equip robots with a set of useful skills for improving learning efficiency~\citep{sutton1999options, schaal2006adaptive,hausman2018learning,lynch2020learning,pertsch2020spirl,haarnoja2023learning}. %
A common approach to acquiring a rich skill set is to pre-train policies on a wide range of tasks.
Recent works have employed \emph{language instructions} as a way for humans to manually define such tasks for policy training, typically via hindsight annotation of large, pre-collected robot experience datasets~\citep{mees2022calvin, lynch2021language,lynch2022interactive,brohan2022rt}. While the resulting policies show impressive capabilities, generalization to new tasks requires a \emph{large} set of pre-trained skills and thus many pre-training tasks. As a result, prior works resorted to annotating robot trajectory datasets with \emph{hundreds of thousands} of human instruction labels~\citep{lynch2022interactive}, limiting their application outside industrial contexts. 
Can we instead devise a pre-training approach that similarly equips robots with a wide repertoire of skills but \emph{minimizes} the need for human task annotations?

We introduce \method\ (\textbf{S}calable \textbf{P}re-training via \textbf{R}elabeling Language \textbf{IN}s\textbf{T}ructions), a scalable pre-training approach that equips robots with a large set of skills while substantially reducing human labeling effort (see Figure~\ref{fig:teaser}). \corlnew{Given an initial set of language-labeled pre-training tasks, \method\ uses extensive \emph{automated} relabeling to greatly expand this task set without additional human effort. }
Given a dataset of robot trajectories with initial language instruction annotations, we leverage two core ideas to grow the number of tasks. First, we leverage the rich knowledge captured in large language models (LLMs) to iteratively combine consecutive language instructions into more complex tasks, \eg ``\textit{place mug in coffee machine}'' and ``\textit{press brew button}'' into ``\textit{make coffee}''. Second, we propose a language-conditioned offline \corlnew{reinforcement learning (RL)} objective that ``stitches'' multiple trajectory segments from the data to form new tasks, a process we call ``skill chaining'' since it allows the policy to learn longer-horizon skills. Through the combination of both techniques, \method\ creates a richer pre-training task set \corlnew{that can help the agent generalize to new tasks}. We demonstrate that \method-pre-trained robots can leverage their resulting larger skill repertoire to more efficiently learn \corlnew{new} downstream tasks.

In summary, our contributions are threefold: (1)~we propose \method, a scalable pre-training approach for robot policies that minimizes human task annotation effort via LLM-based aggregation and cross-trajectory skill chaining, (2)~we introduce ALFRED-RL, an RL benchmark for the popular ALFRED household task simulator~\citep{ALFRED20}, to test our pre-trained agents on a rich set of long-horizon, semantically meaningful tasks, (3) we demonstrate that policies pre-trained with \method\ learn downstream tasks more efficiently than prior pre-training approaches, both on challenging ALFRED tasks and in a real robot kitchen manipulation setup.

\section{Related Work}

\textbf{Language in RL.} 
There is a large body of work at the intersection of natural language processing and behavior learning for robotics, and the field has been further accelerated by the recent successes in training large, general-purpose language models. Language has been used to structure agents' representations~\citep{andreas2017learning,nair2022r3m}, learn reward functions~\citep{fan2022minedojo}, guide task learning via recipes~\citep{branavan2009reinforcement,andreas2016modular} and perform long-horizon planning~\citep{huang2022language,saycan2022arxiv,huang2022inner,singh2022progprompt}. Another line of work has used language to define a wide range of tasks for pre-training policies, resulting in impressive generalization capabilities~\citep{lynch2021language,lynch2022interactive,brohan2022rt}. Yet, these works require collecting hundreds of thousands of costly human language instructions. Our approach \method\ builds on this line of work but introduces two novel objectives for \textit{automatic relabeling} of training task instructions, thereby substantially reducing the amount of human labeling required for successful pre-training. Prior works have also investigated automated language instruction generation~\citep{colas2020language, cideron2020higher, li2022pretrained}, but they focus on online learning and make assumptions that are hard to scale, \eg hand-defined grammars~\citep{colas2020language} or privileged state information~\citep{li2022pretrained,cideron2020higher}. In contrast, we perform \emph{offline} pre-training and use large language models for \emph{scalable} task generation.

\textbf{Pre-training Policies for RL.} 
Developing policy pre-training approaches for faster downstream learning has been investigated for many years~\citep{ijspeert2002movement,theodorou2010reinforcement,hester2018deep}. Recent advances in offline RL ~\citep{levine2020offline} enabled approaches that can pre-train agents offline and effectively finetune them on online tasks~\citep{awr, singh2020cog, nair2020accelerating, kostrikov2022offline}. However, these approaches require target-task reward annotations on the pre-training data and the resulting policies are only pre-trained to solve the target task. Meta-RL approaches, on the other hand, pre-train on a range of tasks and thus allow fast adaptation to \emph{unseen} downstream tasks~\citep{duan2016rl, finn2017model, rakelly2019efficient, nam2022simpl}, yet require the tedious manual definition of pre-training tasks by experts. To avoid manual task design, other works have explored unsupervised pre-training approaches based on behavior diversification~\citep{achiam2018variational,eysenbach2018diversity,Sharma2019}, extraction of behavior priors from offline agent experience~\citep{pertsch2020spirl,ajay2020opal,singh2020parrot} or goal state reaching~\citep{lexa2021, actionablemodels2021arxiv}. 
\corlnew{Closest to ours, \citet{actionablemodels2021arxiv} proposes an objective that randomly selects states to chain together existing trajectories, while we propose a language skill chaining objective that allows \method\ to 
execute new, composite language instructions.}
Such unsupervised pre-training approaches~\citep{actionablemodels2021arxiv} learn skill repertoires without clear meaning, which, as we demonstrate in Section~\ref{sec:experiments}, lead to worse downstream task transfer.%

\textbf{Pre-trained Models for Data Augmentation.} 
Obtaining robot (pre-)training data at scale is costly. Thus, recent works have explored using world knowledge captured in large pre-trained models for enriching robot learning datasets, \eg by increasing the visual diversity of trajectories~\citep{yu2023scaling,chen2023genaug, mandi2023cacti} or annotating unlabeled data~\citep{xiao2022robotic}. Our approach similarly leverages pre-trained (language) models for automated data augmentation. By investigating an orthogonal augmentation direction, aggregation and chaining of natural language instructions, \method\ is complementary to these methods.

\begin{figure}[t]
    \centering
    \includegraphics[width=\linewidth]{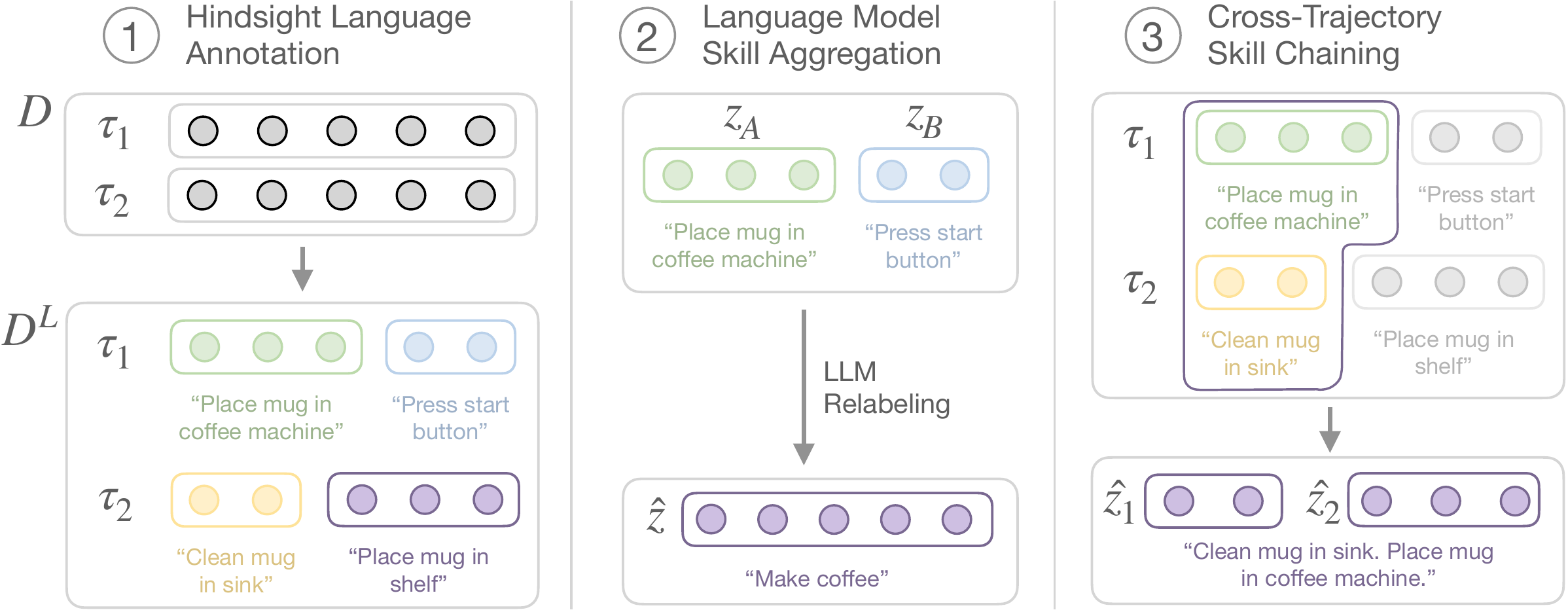}
    \caption{\method\ overview. We assume access to a dataset of agent experience with language instructions for the performed skills \textbf{(1)}. Collecting such instructions with human hindsight annotation is a flexible yet costly approach for defining pre-training tasks. Thus, \method\ introduces two approaches for automatically growing the set of pre-training tasks without additional human effort: \textbf{(2)}~by aggregating language instructions with an LLM and adding the relabeled trajectories back into the pre-training dataset (Section~\ref{sec:llm_relabeling}), \textbf{(3)}~by performing cross-trajectory chaining of skills to enable pre-training of skills that are unseen in the offline agent experience (Section~\ref{sec:cross_chaining}). %
    }
    \label{fig:method}
\end{figure}
\section{\method: Scalable Policy Pre-Training with Language Instructions}
\label{sec:approach}

In this paper, we propose \method~(\textbf{S}calable \textbf{P}re-training via \textbf{R}elabeling Language \textbf{IN}s\textbf{T}ructions), an approach for pre-training robot policies that equips them with a rich repertoire of skills to enable efficient finetuning on unseen tasks. %
Following prior work on agent pre-training, \method\ assumes access to a large offline dataset $\mathcal{D}$ of agent experience~\citep{gupta2019relay,lynch2020learning,pertsch2020spirl,actionablemodels2021arxiv,ebert2022bridge,pertsch2021skild}, collected, \eg from prior RL runs or via teleoperation. We further assume that the data is annotated with an initial set of natural language task instructions, \eg ``\textit{put a mug in the coffee machine}'' or ``\textit{push the brew button}'', that can be collected \emph{in hindsight} via platforms like Amazon Mechanical Turk~\citep{lynch2021language,ALFRED20}. %
Given a sequence $\tau$ of states and actions from the dataset~$\mathcal{D}$, annotators can label sub-trajectories $\tau_1 = [s_0, a_0, s_1, \dots], \tau_2 = \dots$ with free-form language descriptions $z_1, z_2, \dots$ of the skills executed in the respective sub-trajectories (see Figure~\ref{fig:method}, left), resulting in a \emph{language-annotated} dataset~$\mathcal{D}^L$. %

\textbf{Approach Overview.} \method\ equips policies with a diverse repertoire of skills via language-instruction-conditioned offline RL: given a natural language task description $z$, the policy $\pi(a \vert s, z)$ is rewarded for successfully executing the instruction (Section~\ref{sec:semantic_pretrain}). %
Intuitively, the richer the set of task instructions during pre-training, the more skills the policy will learn and the more downstream tasks it can finetune on efficiently. Thus, \method\ introduces two approaches for increasing the scale and diversity of the pre-training task instructions without requiring additional costly human inputs. Firstly, \method\ leverages pre-trained language models to aggregate consecutive instructions into new tasks (Figure~\ref{fig:method}, middle, Section~\ref{sec:llm_relabeling}). 
Secondly, \method\ introduces an objective for cross-trajectory skill-chaining via offline RL that generates novel instruction chains \emph{across different trajectories} (Figure~\ref{fig:method}, right, Section~\ref{sec:cross_chaining}). \method\ pre-trains policies on the combined set of tasks and thereby equips them with a richer skill repertoire. In our experiments (Section~\ref{sec:experiments}) we demonstrate that this leads to more effective learning of new tasks.

\subsection{Instruction-Conditioned Offline RL}
\label{sec:semantic_pretrain}

To pre-train our policy $\pi$ with the natural language instruction dataset $\mathcal{D}^L$, we take inspiration from goal-conditioned RL~\citep{Kaelbling93learningto, universal_value_func, actionablemodels2021arxiv}:
instead of rewarding the policy for reaching goal states, we condition our policy $\pi(a \vert s, z)$ on \emph{language instructions} $z$ from $\mathcal{D}^L$ and provide a scalable sparse reward $R(s, a, z)$ to the agent for reaching the end-state $s_{T}$ of the sub-trajectory. Formally, we define the reward as:
\begin{equation}
    R(s, a, z) =
  \begin{cases}
    1, & \text{for } s = s_{T} \\
    0, & \text{otherwise.}
  \end{cases}
\label{eq:sparse_reward}
\end{equation}
We train our policy $\pi(a \vert s, z)$ to maximize this reward with offline RL~\citep{levine2020offline} using an instruction-conditioned critic $Q(s, a, z)$. Specifically, we use Implicit Q-Learning~\citep{kostrikov2022offline} as it is performant and easy to tune. %

\subsection{Language-Model-Based Instruction Aggregation}
\label{sec:llm_relabeling}
\begin{wrapfigure}[15]{L}{0.5\linewidth}
\vspace{-0.58cm}
\footnotesize
\begin{mdframed}[frametitle=LLM Prompt Example, frametitlealignment=\centering,]
Summarize the following steps.
\\

1: Pick up the tomato slice.

2: Heat it up in the microwave.

Summary: Microwave a tomato slice. 
\\

1: [SKILL 1]

2: [SKILL 2]

...

Summary:
\end{mdframed}
\vspace{-9pt}
\caption{A shortened example of the LLM prompt.
See the full prompt in 
appendix, \mysec{sec:appendix:prompt}.}
\label{fig:short llm prompt}
\end{wrapfigure}
Large language models (LLMs), trained on massive corpora of internet text data, have been shown to be effective at performing a variety of tasks -- from question answering to program synthesis -- when prompted with relevant text~\citep{BERT, GPT3, gpt-j, GOPHER, Chinchilla, OPT, PaLM}. 
Here we use LLMs to \textit{aggregate}, \ie paraphrase, the existing language instructions in $\mathcal{D}^L$ (see Figure~\ref{fig:method}, middle). Given a trajectory that contains multiple sub-trajectories, we can aggregate adjacent sub-trajectories into a longer trajectory and relabel its natural language annotation with a summary of the individual instructions generated by the LLM, thereby generating a new \textit{higher-level} pre-training task that encompasses instructions from multiple sub-trajectories.\footnote{Other relabeling operations, such as splitting an instruction into lower-level instructions, can also be performed by the LLM. However, such operations require grounding the LLM in the agent's observations to determine sub-trajectory split points. We leave investigating this to future work.} We use a simple summarization prompt to instruct the language model (see Figure~\ref{fig:short llm prompt}). Specifically, we aggregate with LLAMA-13B~\citep{touvron2023llama}, an open-source 13 billion parameter LLM \corlnew{which is able to retain important information from individual instructions in the overall summary} (see appendix, Section \ref{sec:appendix:experiments:summaries} for qualitative examples). %
Like in Section~\ref{sec:semantic_pretrain}, the reward for this new aggregated sub-trajectory is 1 at the last transition and 0 otherwise. %
For example, we prompt the LLM to summarize the two skills ($z_1:$ ``\textit{Put a mug in the coffee machine},'' $z_2:$ ``\textit{Push the brew button}''), resulting in a new annotation $\hat{z}_{1:2}$ describing both skills (\eg ``\textit{Make coffee}''). We then add the new trajectory back to our dataset $\mathcal{D}^L$. \corlnew{Using this technique, we generate new language annotations for all combinations of consecutive sub-trajectories in our dataset.} In practice, this increases the number of task instructions by 2.5x in ALFRED and 2x in our robot manipulation dataset (see Section~\ref{sec:experiments}).

\subsection{Cross-Trajectory Chaining}
\label{sec:cross_chaining}
In addition to generating new pre-training tasks composed of behaviors within the \emph{same} trajectory (Section~\ref{sec:llm_relabeling}), we also want to be able to generate pre-training tasks containing behaviors across \emph{different} trajectories.
For example, if trajectory~(A) shows cleaning the mug in the sink %
while trajectory~(B) starts with placing the mug in the coffee machine, the agent should be able to learn to clean the mug in the sink and then place it in the coffee machine (see Figure~\ref{fig:method}, right), thus learning long-horizon behaviors that are unseen in the training data. 
Agents trained with standard offline RL can implicitly combine tasks described from multiple trajectories into longer-horizon behaviors via value propagation, \ie perform ``stitching''~\citep{levine2020offline}.
In our case of \emph{instruction-conditioned} offline RL, values do not naturally propagate from trajectory (B) back to trajectory (A) due to the different language instruction conditionings for the critic $Q(s, a, z_A)$ and $Q(s, a, z_B)$. However, we can actively add ``chaining examples''~\citep{actionablemodels2021arxiv}, which encourage learning longer-horizon behaviors, to our training dataset by first \emph{combining language instructions} and then appropriately \emph{relabeling rewards}.
To build such chaining examples, we first sample two sub-trajectories 
$\tau_{z_A}$ 
and $\tau_{z_B}$ 
from \emph{different} trajectories (see Figure~\ref{fig:method}, right). Next, we create an aggregate instruction $\hat{z}$ which indicates that the agent first finishes (A) and then finishes (B), \eg ``\textit{clean the coffee mug (A) and place it in the coffee machine (B)}.''\footnote{Note that we could generate $\hat{z}$ using the same LLM summarization as in Section~\ref{sec:llm_relabeling}. Yet we found the resulting summaries to often be confusing since randomly paired instructions \emph{from different trajectories} can rarely be summarized meaningfully. We got the best empirical results by simply concatenating the sampled instructions with the word ``and''. Note that we perform chaining on both the original trajectories and those generated by LLM aggregation in Section~\ref{sec:llm_relabeling}.}

Unlike in Section~\ref{sec:llm_relabeling}, we cannot simply concatenate the two trajectories together and relabel the reward of the last transition to 1. Since we sampled the two sub-trajectories at random, the last state of the first, $s_{T_A}$, does not directly transition into the first state of the second.
To solve this issue, we relabel both $\tau_{z_A}$ and $\tau_{z_B}$ with the aggregate instruction $\hat{z}$ and treat them as \emph{separate} trajectories with appropriately labeled rewards. 
For transitions in $\tau_{z_B}$, we simply relabel the last transition with a reward of 1 to be consistent with the 0-1 rewards in Sections~\ref{sec:semantic_pretrain} and \ref{sec:llm_relabeling}. 
Meanwhile, we would like to relabel the reward of the last, terminal transition in $\tau_{z_A}$ so that the learned Q-value for this transition, $Q(s_{T_A}, a_{T_A}, \hat{z})$, will also be consistent with the prior labeling schemes. What reward should we use here?

Recall that Q-functions trained for sparse reward (Eq.~\ref{eq:sparse_reward}) intuitively represent a value proportional to the probability of reaching goal state $s_{T_z}$ at time $T$~\citep{actionablemodels2021arxiv, EysenbachContrastiveRL}:
\begin{align}
    Q^\pi(s_t, a_t, z) &= \E [\sum_{t'=t} \gamma^{t'} R(s_{t'}, a_{t'}, z)] \notag \\
    &=  \E\left[\gamma^{T-t} \mathds{1}\left[ s_T = s_{T_z} \right]\right] 
    \propto P^\pi(s_T = s_{T_z} | s_t, a_t).
    \label{eq:q_probability}
\end{align}
where $\gamma \in (0, 1)$ denotes the discount factor. 
Following this intuition, the Q-value learned for the last
transition of (A) should be proportional to the probability of finishing \emph{the remainder} of the combined task $\hat{z}$, \ie proportional to the likelihood of finishing
(B) from $s_{T_A}$ when taking action $a_{T_A}$. 
Following Eq.~\ref{eq:q_probability}, $Q(s_{T_A}, a_{T_A}, z_B)$ \emph{is} this probability. Intuitively, if there are transitions in the dataset which indicate that finishing (B) from $s_{T_A}$ by taking action $a_{T_A}$ is possible, then this Q-value should be non-zero and the agent will learn to chain (A) and (B) together through their aggregate instruction $\hat{z}$. Our reward labels for the two trajectories with aggregate instruction $\hat{z}$ are therefore:
\begin{equation}
    R(s, a, \hat{z}) =
  \begin{cases}
    1, & \text{for } s = s_{T_B} \\
    Q(s, a, z_B), & \text{for } s = s_{T_A} \\
    0, & \text{otherwise.}
  \end{cases}
\label{eq:cross_chaining_reward}
\end{equation}
Since $Q$ changes during training, we compute the rewards in Eq.~\ref{eq:cross_chaining_reward} in each batch while training. 
For a discussion how chaining preserves the structure of the original MDP, see Appendix Section~\ref{sec:appendix:cross-traj chaining discussion}. 
Full \method\ pseudocode is listed in Alg.~\ref{alg:approach_summary}.
\begin{algorithm}
\caption{\method\ Algorithm} \label{alg:approach_summary}
\begin{algorithmic}[1]
\Require Dataset $\mathcal{D}^L$ w/ language instruction labels, LLM %

\State \textsc{AggregateSkills}($\mathcal{D}^L$, LLM)
\While{not converged}
\State $\tau_z \leftarrow \mathcal{D}^L$: Sample an annotated skill (sub-)trajectory
\State Train offline RL on $\tau_z$
\State $\tau_{\text{agg}_1}, \tau_{\text{agg}_2} \leftarrow $ \textsc{ChainSkills}($\mathcal{D}^L$, LLM)
\State Train offline RL on $\tau_{\text{agg}_1}, \tau_{\text{agg}_2} $
\EndWhile

\Procedure{AggregateSkills}{$\mathcal{D}^L$, LLM}\Comment{Sec.~\ref{sec:llm_relabeling}}
\For{composite trajectory $\tau_{\bar{z}}$ in $\mathcal{D}^L$}
    \For{all adjacent sub-trajectories $\left[\tau_{z_i}...\tau_{z_j}\right]$}
        \State Assign name from LLM: $\text{LLM}(z_i...z_j) = \hat{z}_{i:j}$
        \State $\tau_{\hat{z}_{i:j}} \leftarrow $ Concat $\left[\tau_{z_i},...,\tau_{z_j}\right]$ and relabel with $\hat{z}_{i:j}$ and reward from \myeq{eq:sparse_reward}.
        \State $\mathcal{D}^L = \mathcal{D}^L \, \cup \, \left\{\tau_{\hat{z}_{i:j}}\right\}$
    \EndFor
\EndFor
\EndProcedure

\Procedure{ChainSkills}{$\mathcal{D}^L$, LLM} \Comment{Sec.~\ref{sec:cross_chaining}}
\State Sample random $\tau_{z_1}, \tau_{z_2} \sim \mathcal{D}^L$ 
\State Assign new name : $\hat{z} = $ ``$\{z_1\} \text{ and } \{z_2\}$''
\State $\tau_{\text{agg}_1} \leftarrow$ Relabel $\tau_{z_1}$ w/ $\hat{z}$ and rew from \myeq{eq:cross_chaining_reward}
\State $\tau_{\text{agg}_2} \leftarrow$ Relabel $\tau_{z_2}$ w/ $\hat{z}$ and rew from \myeq{eq:cross_chaining_reward}
\State \Return $\tau_{\text{agg}_1}, \tau_{\text{agg}_2}$
\EndProcedure

\end{algorithmic}
\end{algorithm}
\section{Experiments}
\label{sec:experiments}
In our experiments, we investigate how well an agent pre-trained with \method\ performs on challenging unseen tasks. 
Thus, we answer the following questions: (1)~Does \method\ enable more efficient finetuning on unseen target tasks than previous pre-training approaches? (2)~Can \method\ agents execute unseen language instructions zero-shot? (3)~Does augmentation via \emph{language} relabeling lead to more generalizable policies than through goal image relabeling?

\subsection{Experimental Setup}
\label{sec:setup}

\begin{figure}[h]
\centering
\hfill
\begin{subfigure}[b]{0.46\linewidth}
    \centering
    \includegraphics[width=\textwidth]{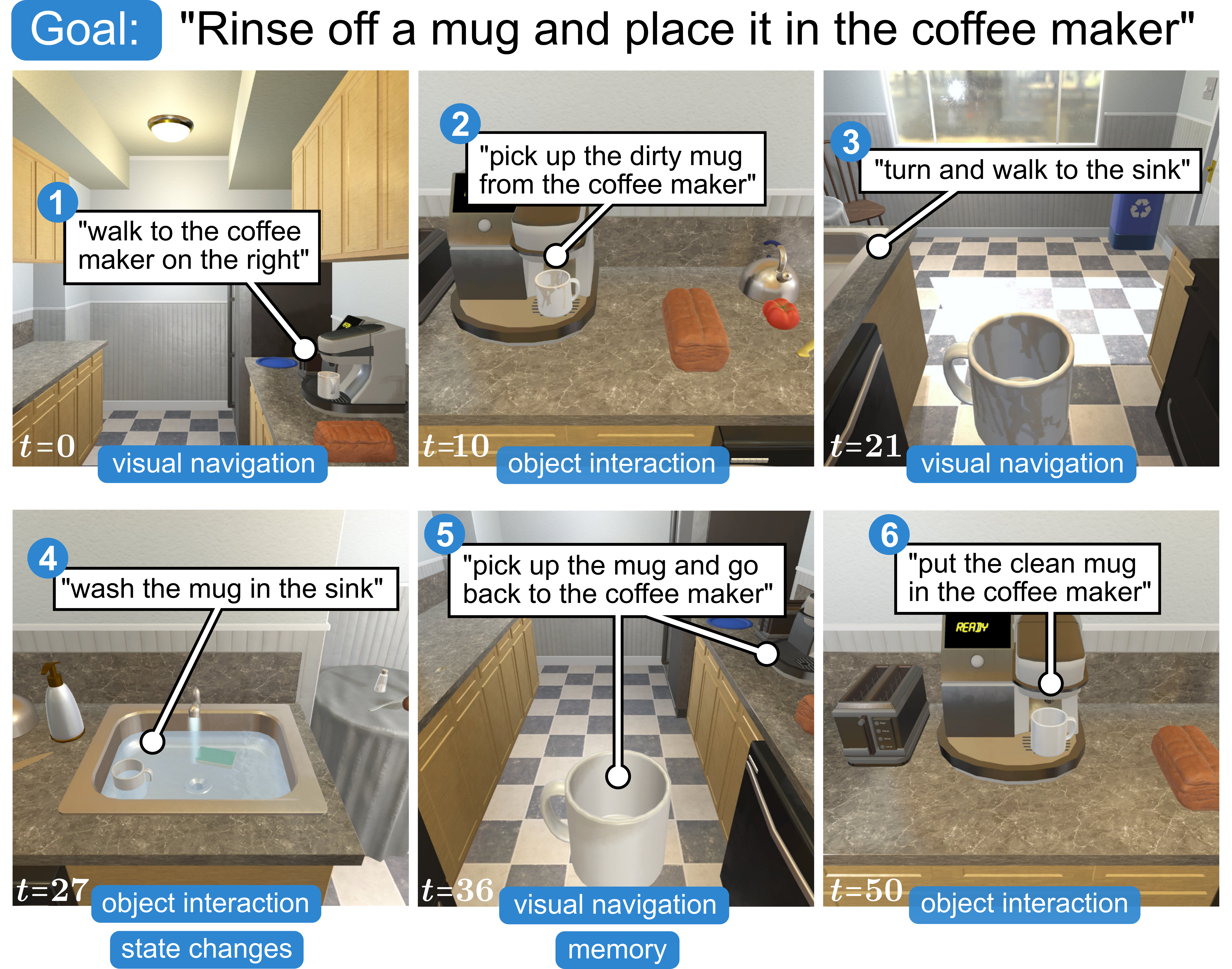}
    \label{fig:alfred}
\end{subfigure}
\hfill
\begin{subfigure}[b]{0.49\linewidth}
\centering
   \includegraphics[width=\linewidth]{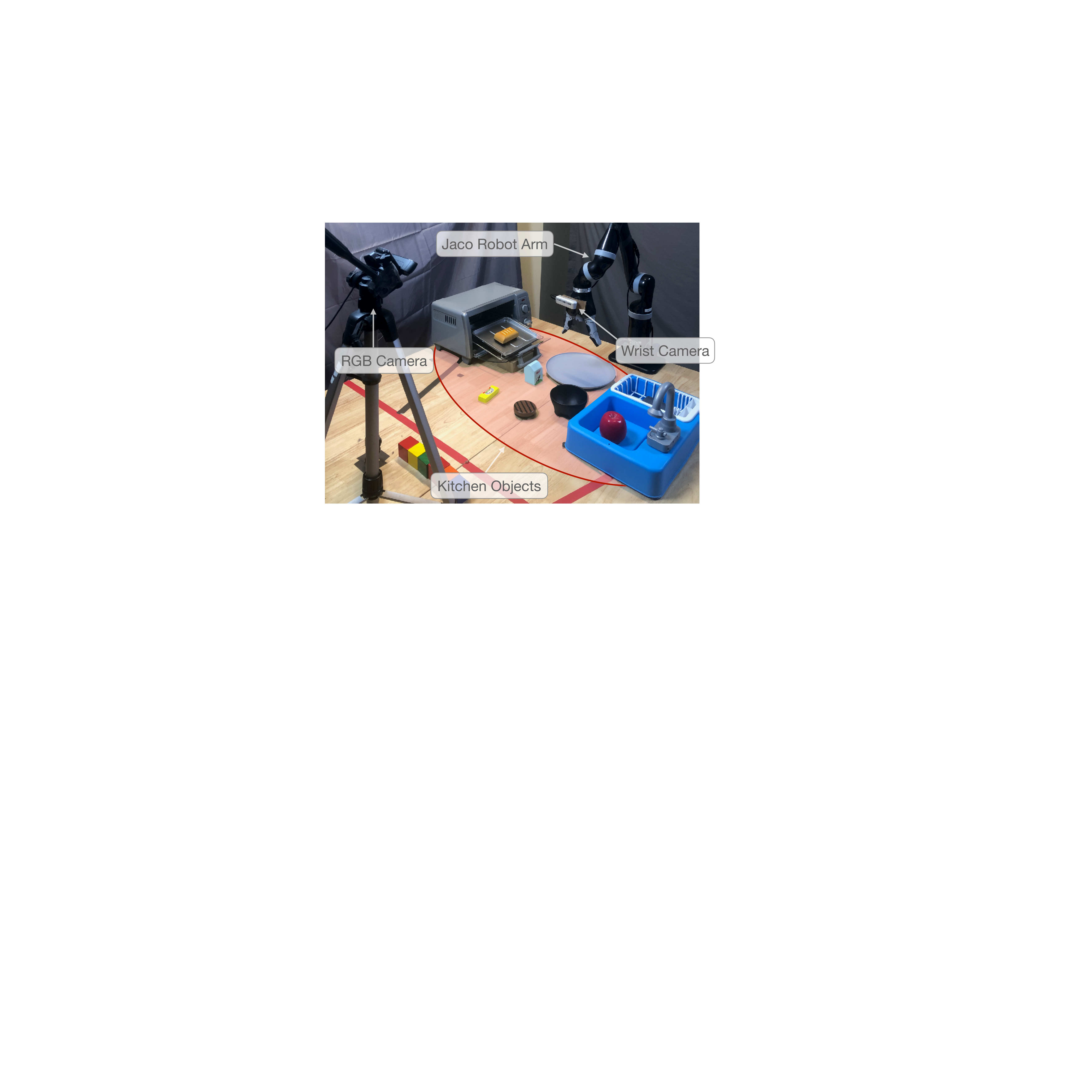} 
   \label{fig:real_robot}
\end{subfigure}
\hfill
\caption{\textbf{Left}: ALFRED provides a rich set of long-horizon, meaningful tasks and a dataset of 6.6k language-annotated demos. We introduce the ALFRED-RL Benchmark which tests finetuning of RL agents on unseen tasks and scenes. \textbf{Right}: Our Jaco robot arm with RGB image-based control.} %
\label{fig:environments}
\end{figure}

We evaluate our approach on two image-based environments (see Figure~\ref{fig:environments}): ALFRED-RL, an RL benchmark we develop upon \citet{ALFRED20}, and a real robot kitchen with a Jaco 2 robot arm.

\textbf{ALFRED-RL.} Our goal is to compare different pre-training approaches on a diverse set of semantically meaningful, long-horizon tasks. Yet, existing multi-task RL environments typically evaluate only on short-horizon or semantically meaningless tasks~\citep{yu2019metaworld,mees2022calvin}. Thus, we introduce a new RL benchmark based on the ALFRED household task simulator~\citep{ALFRED20}. While ALFRED abstracts away low-level agent control into discrete actions like ``pick up'' or ``turn left,'' its 100+ rich indoor scenes with many interactable objects allow to evaluate an agent's capabilities for solving long-horizon household tasks from a rich task distribution. The original benchmark focuses on imitation learning, but we extend it to support training RL agents through a gym interface with egocentric RGB observations and an action space consisting of 12 discrete 
action choices and 82 interactable object types~\citep{pashevich2021episodic}. We create three evaluation task sets that test progressively more challenging axes of generalization: \textbf{\emph{EVAL\textsubscript{INSTRUCT}}} uses unseen human-generated instructions on familiar scenes, \textbf{\emph{EVAL\textsubscript{LENGTH}}} uses tasks that are longer than any observed in pre-training, testing ``stitching'' capabilities, and \textbf{\emph{EVAL\textsubscript{SCENE}}} uses tasks in unseen floorplans. 
For more details about ALFRED-RL and evaluation set construction, see Appendix, Section~\ref{sec:appendix:eval_dataset}.

\textbf{Real-World Robot Kitchen Manipulation.} 
To evaluate pre-training approaches on end-to-end \emph{low-level} robot control, we design a set of stylized kitchen manipulation tasks with a Kinova Jaco 2 robot arm. The policy's inputs are RGB images from a wrist-mounted and a third-person camera and it produces continuous end-effector (3-dim) displacement actions and a discrete gripper open/stay/close action at a control frequency of 10Hz. We collect a dataset of 329 long-horizon trajectories via human teleoperation with the setup from \citet{dass2023jacoplay}, each consisting of multiple language-annotated sub-trajectories like \textit{``pick up the apple fruit,''}, \textit{``place the black bowl in the dish rack,''} etc. For evaluation, we construct three long-horizon tasks, sequencing 2 to 8 ``primitive skills'' like the ones mentioned above, in environment configurations that are unseen in the pre-training data. We collect 25 demonstrations for each of the three tasks to evaluate offline fine-tuning performance of different pre-trained policies.

\textbf{Comparisons.} We compare \method\ against common policy pre-training approaches, behavioral cloning and offline goal-conditioned RL: %
\textbf{Language-conditioned BC (L-BC)}~\citep{jang2021bc, lynch2021language}: Behavior cloning (BC) conditioned on the individual language instructions; %
\textbf{Episodic Transformers (ET)}~\citep{pashevich2021episodic}: BC conditioned on sequences of language instructions -- ET is the best-performing end-to-end learned policy on the ALFRED leaderboard that \emph{does not} use privileged domain knowledge like hand-engineered policies or voxel maps\corlnew{;} %
\textbf{Actionable Models (AM)}~\citep{actionablemodels2021arxiv}: Goal-conditioned offline RL with randomly sampled goal observations from the same training data as \method.
\corlnew{We also evaluate \textbf{SayCan}~\citep{saycan2022arxiv}: Top-down LLM planning over pre-trained, language-conditioned policies.}

All methods use the same architectures, hyperparameters, and training data $\mathcal{D}^L$ where possible.
In ALFRED-RL, all methods use the same transformer-based architecture inspired by ~\citet{pashevich2021episodic, snell2023offline}, and on the real robot, they use an RNN architecture with ``action chunking''~\citep{zhao2023learning} first proposed by \citet{dass2023pato}.
For more implementation details, see appendix \mysec{sec:appendix:baselines}. 
All results are means and standard deviations over 3 seeds.

\begin{figure*}[ht]
    \centering
    \includegraphics[trim=10 355 230 280, clip, width=\textwidth]{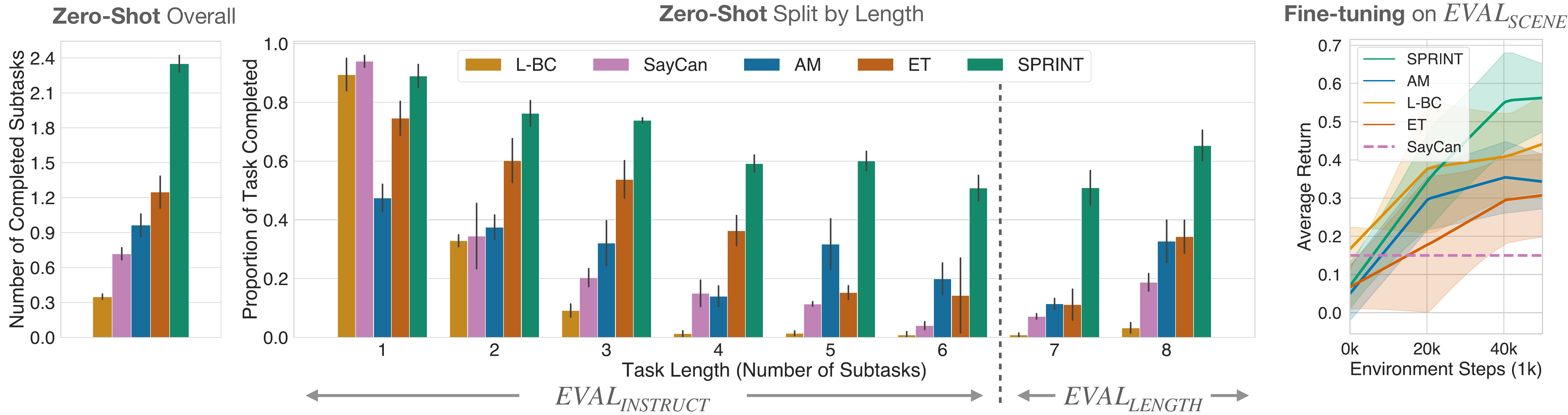}
    \caption{ALFRED-RL evaluation results. \textbf{Left}: Zero shot performance on EVAL\textsubscript{INSTRUCT} and EVAL\textsubscript{LENGTH}. \textcolor{sprint}{SPRINT} is able to complete substantially more subtasks than prior approaches. \textbf{Middle}: Breakdown of performance by task length. \method\ performs well on challenging, long tasks. 
    Numerical results in appendix \mytable{tab:zero_shot_numbers}. 
    \textbf{Right}: Finetuning performance in unseen floor plans of EVAL\textsubscript{SCENE}. \method\ learns in new floorplans more effectively by reaching higher performance.}%
    \label{fig:quant_results}
\end{figure*}

\subsection{\method\ Solves Long-Horizon Tasks Zero-Shot}
\label{sec:zeroshot_results}
We first test the effectiveness of \method's pre-training by analyzing zero-shot performance across 100 unseen tasks in the \textit{EVAL\textsubscript{INSTRUCT}} evaluation set. We report results in Figure~\ref{fig:quant_results}~(left). Our approach, \method, achieves \textbf{2-8x} higher zero-shot task performance than prior pre-training approaches AM and L-BC. Even though ET also trains to condition on long-horizon instruction sequences like \method, ours still outperforms it overall by 2x. To better understand the differences between the methods, we report the breakdown of returns by length of the evaluation task in Figure~\ref{fig:quant_results}~(middle). \corlnew{We find that all methods except AM achieve similar performance on length 1 tasks.} However, on long-horizon tasks, \method\ achieves much higher returns than all baselines since it can leverage the LLM to automatically generate longer-horizon pre-training tasks. In contrast, L-BC trains only on the human-provided, shorter-horizon annotations and thus cannot zero-shot perform longer tasks. \corlnew{Meanwhile SayCan, with the same LLM as used for \method, commonly generates incorrect plans that lead to incorrect behaviors. This problem is exacerbated on longer tasks; the chance of planning errors increases with task length.} %
\corlnew{In contrast, \method's pre-training enables more robust long-horizon task execution.}
Similar to our approach, AM trains to reach long-horizon goals during pre-training but the results in Figure~\ref{fig:quant_results}~(left) show that its pre-training with goal-state conditioning is \emph{less} effective than our language-conditioned pre-training. These results also hold for the \textit{EVAL\textsubscript{LENGTH}} task set, which tests generalization to task horizons beyond the ones seen during training. On these most challenging tasks, \method\ outperforms the best baseline by 2.5x.
(see appendix Figure~\ref{fig:quali_zero_shot} for an example trajectory, appendix Section~\ref{sec:appendix:video_demo} for qualitative comparisons).

\subsection{\method\ Finetunes Effectively in Unseen Environments}
\label{sec:finetuning}
\textbf{ALFRED-RL.}
We test \method's finetuning performance to unseen tasks on the most challenging \textit{EVAL\textsubscript{SCENE}} task set in unseen household floor plans with 50k environment interactions. This corresponds to a realistic scenario in which an agent is placed in a new household environment and needs to leverage skills learned during pre-training to solve new tasks with minimal environment interaction. 
To implement finetuning for \method\ and AM, we condition the policy on a language instruction or goal image from the target task respectively and then run IQL with online data collection. For L-BC and ET, we first pre-train a language-conditioned critic with IQL on the pre-training dataset and then finetune both the policy and critic with online IQL. %
\corlnew{Sparse, per-subtask completion reward is given to agents during fine-tuning.}

We report finetuning results in Figure~\ref{fig:quant_results}~(right). 
with qualitative examples in Appendix, Section~\ref{sec:appendix:video_demo}. 
\method\ quickly achieves higher downstream task return %
than the best prior work. Specifically, L-BC converges to lower peak performance than \method\ and ET performs poorly, perhaps because transferring from instruction sequences to high-level task descriptions is challenging.
Meanwhile, AM performs similarly to L-BC, possibly because unseen goal states are more difficult to learn from.
In contrast, \method's pre-training with language conditioning allows for effective transfer even to unseen environments since the semantics of the tasks transfer well: the language description ``\textit{place cup in coffee machine}'' transfers to many environments while the goal image for the same task might look very different. %
Thus, pre-training with language instructions can enable better transfer for learning tasks in new environments than pre-training to reach goal states. SayCan performs poorly due to both planning and execution errors as it does not fine-tune. We also attempted to first fine-tune SayCan’s primitive policies before running SayCan, but its performance did not change as fine-tuning its policies on high-level task instructions did not improve primitive instruction execution.
          
\textbf{Real Robot.} We also measure finetuning performance on an unseen environment on our real robot setup. We evaluate on three tasks consisting of 2, 4, and 8 subgoals, respectively: 
\begin{enumerate}
    \item \textit{Bake bread in the oven}: The robot must (1) pick up the bread, (2) place it in the oven.
    \item \textit{Serve heated milk in the bowl}: The robot must (1) pick up the milk, (2) place it in the black bowl, (3) pick up the bowl with milk, (4) place the bowl in the oven.
    \item \textit{Serve milk in the bowl and butter and baked bread in the plate}: (1) pick up milk, (2) put it in the black bowl, (3) pick up butter, (4) put it in the plate, (5) pick up the bread, (6) bake it in the oven, (7) pick up the bread from the oven, (8) place the bread in the plate.
\end{enumerate}

We collect 25 demonstrations per task for offline finetuning.
We compare \method\ against \textbf{L-BC}, a version of L-BC trained on full sequences of concatenated language instructions (\textbf{L-BC Composite}), and a method that is trained only on the downstream task demonstrations (\textbf{No pre-train}). %

\begin{figure}[ht]
    \centering
    \includegraphics[width=\linewidth]{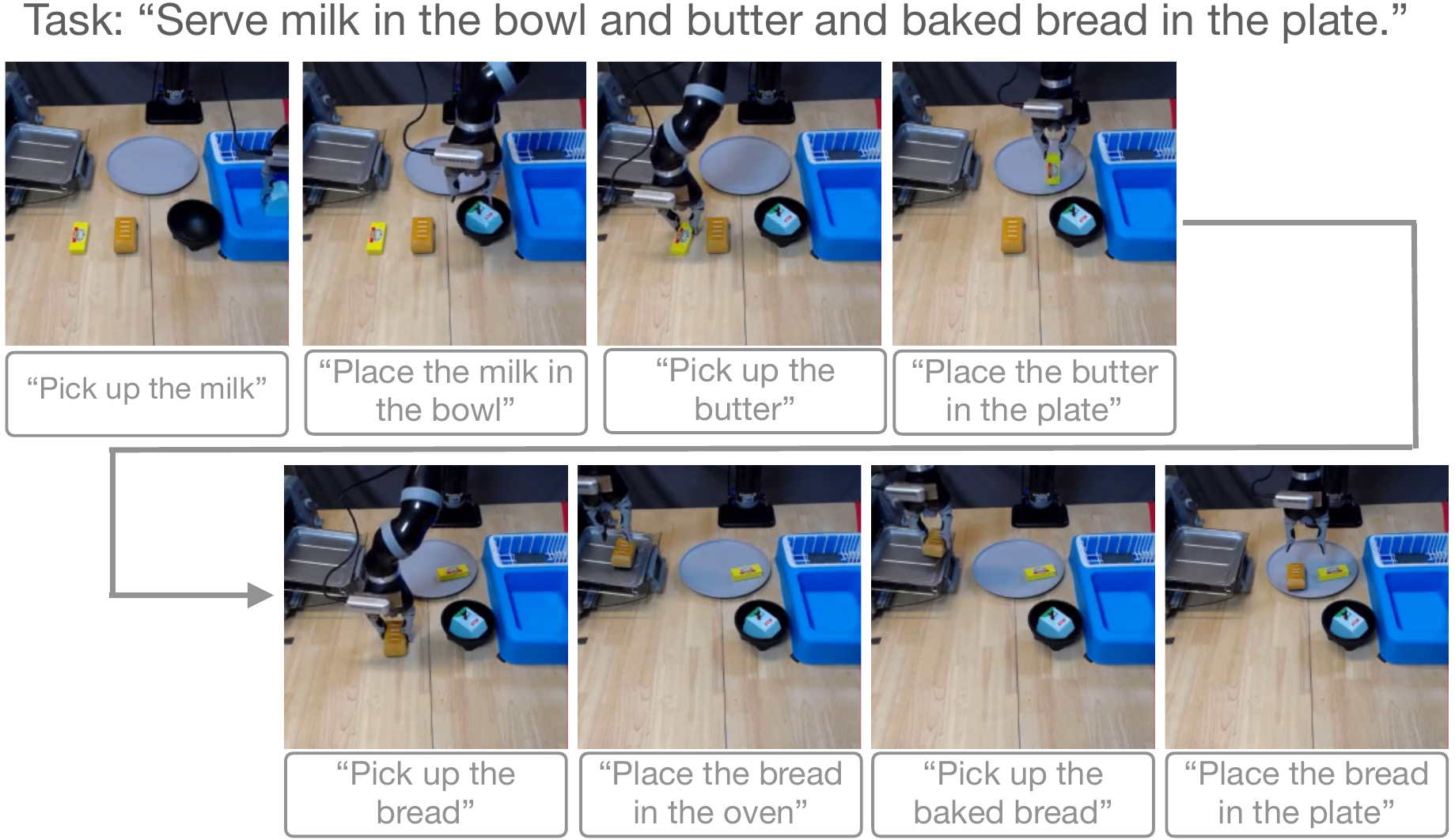}
    \caption{Successful rollout of a \method\ agent offline finetuned for the task above with object combinations not in the pre-training data. \method\ solves all 8 tasks in sequence. }
    \label{fig:quali_real_robot_zero_shot}
\end{figure}

\begin{table}
\caption{Success rates and number of subgoals completed after fine-tuning on the tabletop arrangement displayed on the left with unseen object combinations over 5 trials.}\label{tab:real_robot_finetuning}
\centering
\resizebox{0.9\linewidth}{!}{%
\begin{tabular}{lcccccc}\toprule  
                 & \multicolumn{2}{c}{Length 2} & \multicolumn{2}{c}{Length 4} & \multicolumn{2}{c}{Length 8} \\ 
                 \cmidrule(lr){2-3}\cmidrule(lr){4-5}\cmidrule(lr){6-7}
                 Method & Success & \# Tasks & Success & \# Tasks & Success & \# Tasks \\
                \midrule
                
                \method\ & \textbf{100\%} 
                & \textbf{2.0} & \textbf{60\%} & \textbf{3.4} & \textbf{40\%} & \textbf{6.2}\\%\midrule
                L-BC Comp. & \textbf{100\%} & \textbf{2.0} & 40\% & 2.8 & 20\% & 5.2 \\% \midrule
                L-BC & \textbf{100\%} & \textbf{2.0} & 40\% & 0.4  & 0\% & 2.0 \\% \midrule
                No pre-train & 0\% & 1.0 & 0\% & 0.0 & 0\% & 0.0 \\
                \bottomrule
            \end{tabular}}
\end{table}
Results in Table~\ref{tab:real_robot_finetuning} demonstrate that \textit{No Pre-train} performs poorly, indicating that pre-training is necessary. \method\ achieves the best success rates and completes the most subgoals on all tasks. Compared to L-BC Composite, \method\ achieves higher returns and success rates on challenging, longer tasks. See Figure~\ref{fig:quali_real_robot_zero_shot} for an example evaluation and appendix Section~\ref{sec:appendix:video_demo} for more visualizations.

\subsection{Ablation Studies}
\label{sec:experiments_ablations}
\begin{table}[h]
\small
\caption{%
Ablations. \method\ achieves the highest return with both objectives.}\label{tab:ablation}
\centering
\resizebox{0.80\linewidth}{!}{%
\begin{tabular}{lcc}\toprule  
Ablation &\textit{EVAL\textsubscript{INSTRUCT}} & \textit{EVAL\textsubscript{LENGTH}}\\
\midrule

\method\ (ours) & \textbf{1.94 $\pm$ 0.04} & \textbf{4.40 $\pm$ 0.39}\\%\midrule
\method\ w/o Chain & 1.75 $\pm$ 0.11 & 3.98 $\pm$ 0.29\\%\midrule
\method\ Na\"{i}ve Chain & 0.50 $\pm$ 0.04 & 0.26 $\pm$ 0.05\\
\method\ w/o LLM-agg & 0.37 $\pm$ 0.01 & 0.15 $\pm$ 0.10\\
\bottomrule
\end{tabular}}
\end{table}
We verify the effectiveness of the components of our approach, with the following ablations:
\textbf{\method\ w/o chain} removes cross-trajectory chaining (Section~\ref{sec:cross_chaining}), instead trains only on within-trajectory human-provided and LLM-aggregated tasks; 
\corlnew{\textbf{\method\ Na\"{i}ve Chain} replaces Q-value reward labels when chaining with 0's to test na\"{i}ve offline RL ``stitching'' with language instruction-conditioned agents.}
\textbf{\method\ w/o LLM-agg} additionally removes LLM aggregation (Section~\ref{sec:llm_relabeling}) \corlnew{and chaining}, thus training only on the human-provided task annotations. 
We report zero-shot ALFRED evaluation results in Table~\ref{tab:ablation}: each component of our approach improves zero-shot evaluation performance.
There is a large performance loss when removing LLM aggregation, underlining the importance of leveraging the world knowledge in LLMs for automatically generating long-horizon training tasks. 
\corlnew{We also see that na\"{i}ve chaining \emph{hurts} performance, making it worse than not chaining.}

\section{Discussion and Limitations}
We presented \method, an approach for scalable agent pre-training that automatically generates training tasks for offline RL via LLM relabeling and cross-trajectory skill chaining. We demonstrated that \method\ pre-training leads to higher zero-shot and finetuning performance on diverse household tasks in the ALFRED simulator and on real-robot kitchen manipulation tasks.

\section*{Acknowledgements}
We would like to thank Shivin Dass, Sidhant Kaushik, Laura Smith, Siddarth Verma, and Jullian Yapeter for assisting with task instruction labeling.
We thank Taewook Nam for assisting with the actionable models implementation we based some of our code on, Xiang Ren for detailed feedback on earlier versions of the paper, 
and Yevgen Chebotar for helping us tune the actionable models baseline.
Finally, we thank all of the CLVR lab members at KAIST and USC for their constructive feedback.

This work was supported by a USC Viterbi Fellowship. Additionally, it was supported by Institute of Information \& Communications Technology Planning \& Evaluation (IITP) grants (No.2019-0-00075, Artificial Intelligence Graduate School Program, KAIST; No.2022-0-00077, AI Technology Development for Commonsense Extraction, Reasoning, and Inference from Heterogeneous Data), National Research Foundation of Korea (NRF) grant (NRF-2021H1D3A2A03103683), funded by the Korean government (MSIT), and funded by the KAIST-NAVER hypercreative AI center.

\newpage
\printbibliography
\clearpage
\onecolumn
\appendices  %

\begin{figure*}[h]
    \centering
    \begin{mdframed}
Instructions: give a high-level description for the following steps describing common household tasks.\\

Task Steps:
1. Pick up the keys on the center table.
2. Put the keys in the box.
3. Pick up the box with keys.
4. Put the box with keys on the sofa close to the newspaper.\\
Summary: Put the box with keys on the sofa.\\

Task Steps:
1. Pick up the knife from in front of the tomato.
2. Cut the lettuce on the counter.
3. Set the knife down on the counter in front of the toaster.
4. Pick up a slice of the lettuce from the counter.
5. Put the lettuce slice in the refrigerator. take the lettuce slice out of the refrigerator.
6. Set the lettuce slice on the counter in front of the toaster.\\
Summary: Put a cooled slice of lettuce on the counter.\\

Task Steps:
1. Pick up the book on the table, in front of the chair.
2. Place the book on the left cushion of the couch.\\
Summary: Put a book on the couch.\\

Task Steps:
1. Pick up the fork from the table.
2. Put the fork in the sink and fill the sink with water, then empty the water from the sink and remove the fork.
3. Put the fork in the drawer.\\
Summary: Put the cleaned fork in a drawer.\\

Task Steps:
1. Take the box of tissues from the makeup vanity.
2. Put the tissues on the barred rack.
3. Take the box of tissues from the top of the toilet.
4. Put the tissues on the barred rack.\\
Summary: Put the box of tissues on the barred rack.\\

Task Steps:
1. Pick up the glass from the sink.
2. Heat the glass in the microwave.
3. Put the glass on the wooden rack.\\
Summary: Put a heated glass on the wooden rack.\\

Task Steps:
1. Pick up the box from the far side of the bed.
2. Hold the box and turn on the lamp.\\
Summary: Look at the box under the lamp light.\\

Task Steps:
1: [SKILL 1].
2: [SKILL 2].
3: [SKILL 3].
...
N: [SKILL N].\\
Summary: 
    \end{mdframed}
    \caption{The full prompt that we use for summarization. Following the suggestions of \citet{saycan2022arxiv} for prompt design, we explicitly number each step. The LLM completion task begins after ``Summary:''. For brevity, we omit the new line characters between all numbered steps.}
    \label{fig:llm_prompt}
\end{figure*}

\begin{figure*}
    \centering
    \includegraphics[width=\textwidth]{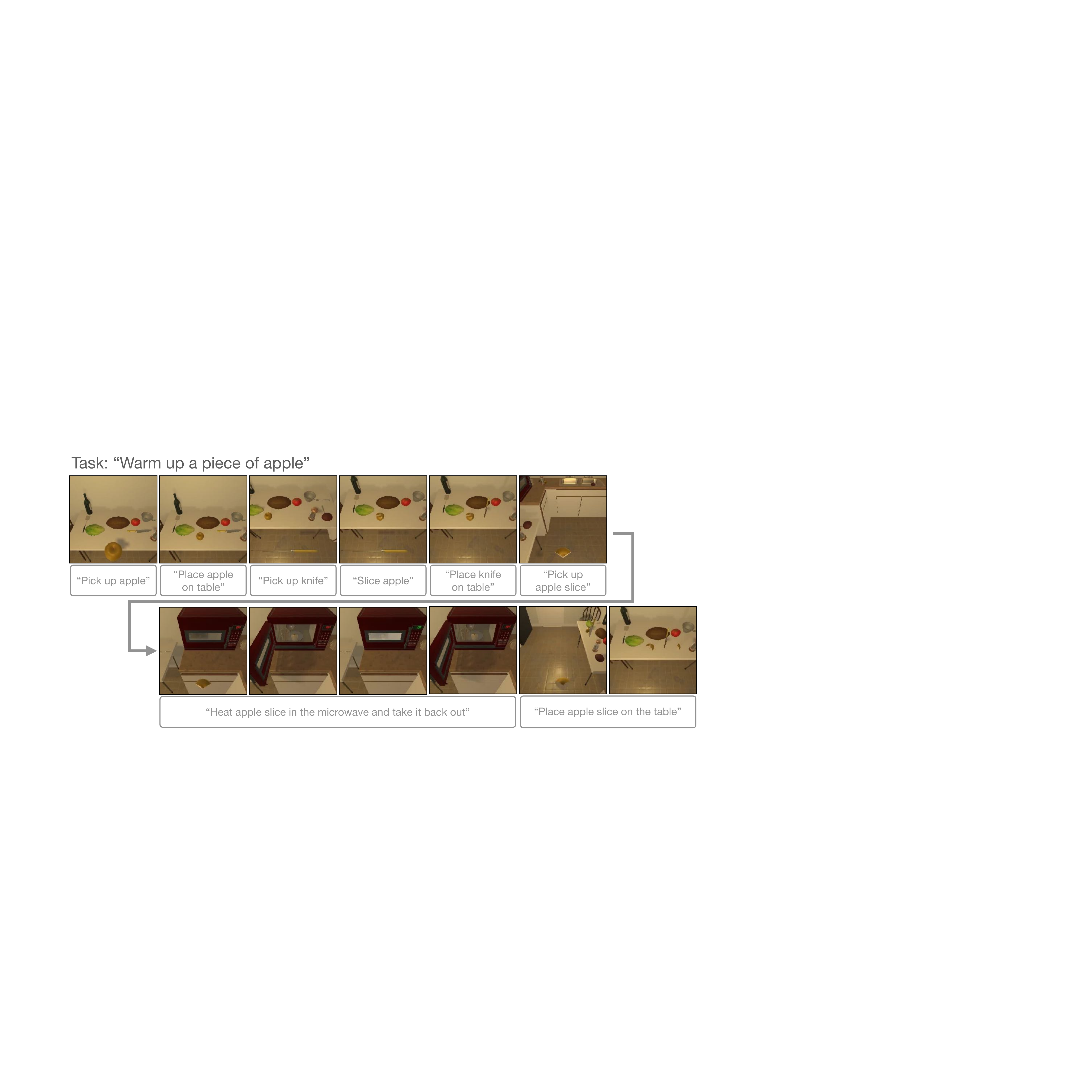}
    \caption{Example successful task execution of our pre-trained \method\ agent for the challenging ``\textit{Warm up a piece of apple}'' task. Successful execution requires solving 8 subtasks in sequence and a total of 50~steps. This sequence of subtasks was never observed in the training data. \method\ uses cross-trajectory stitching and LLM aggregation to learn unseen tasks.}
    \label{fig:quali_zero_shot}
    
\end{figure*}

\section{Large Language Model Prompt}
\label{sec:appendix:prompt}

We list the full large language model summarization prompt in \myfig{fig:llm_prompt}. The examples in the prompt are fixed for all summarization queries. These examples are selected from the ALFRED validation dataset (which is not otherwise used in our work) at random: 
We spell out the primitive skill annotations in the ``Task Steps:'' part of each prompt example. Then, the ``Summary'' for each of these is the high-level, human-written annotation for that trajectory from ALFRED. We repeatedly sampled these trajectories until each example mentioned a different object to prevent biasing the LLM towards certain types of objects. 

We note that the ``Look at the box under the lamp light'' example is important to make the LLM give reasonable summaries for similar tasks in ALFRED where the agent picks something up and turns on a light. This is because most of the human labels for turning on the lamp do not mention the object in the previous step, making it difficult for the LLM to realize that the task has to do with looking at the held object under a lamp.

\section{Baselines and Implementation}
\label{sec:appendix:baselines}
We implement IQL~\citep{kostrikov2022offline} as the base offline RL algorithm for all goal-conditioned offline RL pretraining baselines and ablations due to its strong offline and finetuning performance on a variety of dense and sparse reward environments. At a high level, IQL trains on in-distribution $(s, a, s', r, a')$ tuples from the dataset rather than sampling a policy for $a'$ to ensure that the Q and value functions represent accurate estimated returns constrained to actions in the dataset. The value function is trained with an expectile regression loss controlled by a hyperparameter $\tau$, where $\tau=0.5$ results in standard mean squared error loss and $\tau \rightarrow 1$ approximates the max operator, resulting in a more optimistic value function that can better ``stitch'' together trajectories to obtain distant reward in sparse reward settings. The IQL policy is trained to maximize the following objective: $$e^{\beta \left(Q(s, a) - V(s)\right)}\log \pi(a|s), $$
which performs advantage-weighted regression~\citep{awr} with an inverse temperature term $\beta$. In practice, the exponential advantage term is limited to a maximum value to avoid numerical overflow issues.
We detail shared training and implementation details below, with method-specific information and hyperparameters in the following subsections.

\subsection{ALFRED Details}
\textbf{Observation space.}
The state space of the ALFRED environment consists of 
$300 \times 300$ RGB images. Following the baseline method in ALFRED~\citep{ALFRED20}, we 
preprocess these images by sending them through a frozen ResNet-18 encoder~\citep{he2016deep} pretrained on ImageNet~\citep{imagenet}. This results in a $512 \times 7 \times 7$ feature map that we use
as the observation input to all networks. 

\textbf{Action space.} %
The agent chooses from 12 discrete low-level actions. There are 5 navigation actions: \texttt{MoveAhead}, \texttt{RotateRight}, \texttt{RotateLeft}, \texttt{LookUp}, and \texttt{LookDown} and 7 interaction actions: \texttt{Pickup}, \texttt{Put}, \texttt{Open}, \texttt{Close}, \texttt{ToggleOn}, \texttt{ToggleOff}, and \texttt{Slice}. 
For interaction actions the agent additionally selects one of 82 object types to interact with, as defined by \citet{pashevich2021episodic}.
In total, the action space consists of $5 + 7 * 82= 579$ discrete action choices. Note that this action space definition is different from the action space in \citet{ALFRED20}, which used a pixel-wise mask output to determine the object to interact with. In contrast to \citet{ALFRED20} we aim to train agents with \emph{reinforcement learning} instead of imitation learning and found the discrete action parametrization more amenable to RL training than dense mask outputs. For all methods, due to the large discrete action space, we perform some basic action masking to prevent agents from taking actions that are not possible. For example, we do not allow the agent to \texttt{Close} objects that aren't closeable nor can they \texttt{ToggleOn} objects that can't be turned on.

\textbf{Policy and critic networks.}
For all baselines and \method\, base models are implemented on the transformer architecture proposed
in Episodic Transformers~\citep{pashevich2021episodic}. For offline RL methods (AM, \method) we follow the advice of~\citep{snell2023offline} and parameterize both Q functions and the Value function of IQL as separate output heads of one transformer backbone that is used for all critic networks.
We train both policies and critic transformer networks with an observation history of up to 16 previous observations, each one being processed by a convolutional network before being flattened into a 768-dim feature.
Our discrete policy has two output heads of size 12 and 82 for the action and interaction object outputs respectively. Critic networks are conditioned on both the observation and the discrete action output of the policy. In networks with language input, words are individually tokenized and the entire language instruction is fed to the policy and critic networks and embedded into a sequence of learned 768-dim embeddings, one for each token. We perform cross-attention between all network inputs: language embeddings, previous observation embeddings, and the previous action where applicable.  The output of this cross-attention mechanism is then transformed by linear layers into the final output for the network. %

\textbf{Pre-training hyperparameters.}
A hyperparameter search was performed first on the language-conditioned BC-baseline to optimize for training accuracy. These hyperparameters were carried over to the IQL implementation, and another search for IQL-specific hyperpameters were performed on a baseline IQL policy conditioned on language instructions. With these parameters fixed, we performed one more hyperparameter search specific to Actionable Models but for the final implementation of \method\ we re-used the same hyperparameters and only selected \method-specific parameters heuristically.
Hyperparameters for each method are detailed in separate tables. Shared hyperparameters for all methods (where applicable) are listed below:

\begin{center}
\begin{tabular}{ll}
    \toprule
    Param & Value\\
    \midrule
    Batch Size & 1024\\
    \new{\# Training Batches} & 140k \\
    Learning Rate & 1e-4\\
    Optimizer & AdamW\\
    Dropout Rate & 0.1\\
    Weight Decay & 0.1 \\
    Discount $\gamma$ & 0.97 \\
    Q Update Polyak Averaging Coefficient & 0.005\\
    Policy and Q Update Period & 1/train iter\\
    Nonlinearity & ReLU \\
    IQL Advantage Clipping & [0, 100]\\
    IQL Advantage Inverse Temperature $\beta$ & 5\\
    IQL Quantile $\tau$ & 0.8\\
    Maximum Transformer Context Length & 16\\
    \bottomrule
\end{tabular}
\end{center}

\textbf{Finetuning details and hyperparameters.}
We fine-tune by running IQL on online-collected data without any of the chaining or aggregation steps.
For all models, we finetune by sampling old pre-training data and newly collected data at a ratio of 70\%/30\%. Without this mixed batch training, we found the transformer-based networks to overfit to the new data, something we did not see when experimenting with standard MLPs.
The newly collected data is also sampled from two separate buffers at equal proportions, one which contains trajectories that received at least 1 reward (\ie completed one sub-task) and one that contains trajectories that received none. This is another transformer-specific adaptation we had to make for the models to train stably with IQL on online-collected data.

Each method is finetuned on every task in the \textit{EVAL\textsubscript{SCENE}} task set individually; that is, 
we pre-train once and then finetune policies for each task in the task set. We then average
returns over all tasks, then report metrics averaged over all random seeds. For each task, we define a maximum rollout time horizon of 2 timesteps per environment action required by an expert ALFRED task planner.

When not specified, finetuning parameters are identical to pre-training parameters. Finetuning hyperparameters
are specified below: 

\begin{center}
\scalebox{1.0}{\begin{tabular}{ll}
    \toprule
    Param & Value\\
    \midrule
    \# Initial Rollouts & 50\\
    Training to Env Step Ratio & 20\\
    $\epsilon$ in $\epsilon$-greedy action sampling & 0\\
    Policy action sampling & True\\
    \# Parallel Rollout Samplers & 10\\
    \bottomrule
\end{tabular}}
\end{center}

\subsection{Real Robot Implementation Details}
The real-world environment uses a Kinova Jaco 2 robot arm. Below we detail the implementation and training details specific to the real robot environment.

\paragraph{Observation space.}
The view observations consist of 224×224×3 cropped RGB images, which are captured from a Logitech Pro Webcam C920 for the third-person view and an Intel RealSense D435 for the wrist-view. We leverage a pretrained R3M~\citep{nair2022r3m} model to encode each view observation. Additionally, the state representation includes the robot's end-effector position, velocity, and gripper state. Notably, the end-effector position and velocity are two continuous vectors, while the gripper state is represented as a one-hot vector, indicating OPEN, CLOSE, or NOT MOVE. To form the observation for the policy, we concatenate the embedded RGB input with state information. 

To condition on language inputs, we use a pre-trained sentence embedder to embed the entire language annotation into a vector of size 384 (as our network backbone is an RNN instead of a transformer). This embedding is done with the \texttt{all-MiniLM-L12-v2} pre-trained embedding model from the \texttt{SentenceTransformers} package~\citep{reimers-2019-sentence-bert}. 

The total state input dimension is: 2048 (third-person R3M) + 2048 (wrist R3M) + 15 (Jaco state input) + 384 (language embedding) = 4495.

\paragraph{Action space.}
The robot action space comprises the changes in the end effector position between each time stamp, along with the gripper opening/closing commands. These actions are transmitted to the robot at a frequency of 10 Hz and interpreted as desired joint poses using PyBullet's inverse kinematics module.

\paragraph{Network architecture and training.} 
Similar to ~\citep{zhao2023learning}, we use the Action Chunking method to train an autoregressive policy. Specifically, our policy employs an LSTM model to predict the next 15 actions, given the initial observation as input, i.e., \textit{$\pi$\rm{(}$a_{t:t+15}$$\mid$$s_t$\rm{)}}. Our Q and Value networks are also recurrent, predicting per-timestep rewards for each action in the sequence. Just like the policy, they also only see the observation before the action sequence starts.

Because of the fact that the gripper action is discrete and heavily imbalanced in class distribution, we weigh the gripper action loss inversely proportionally to the number of examples in each class.

\paragraph{Pre-training details and hyperparameters.}
We performed a heuristic hyperparameter search by first tuning the language-conditioned BC baseline to be as effective as possible on zero-shot evaluations of training tasks, then performed a small heuristic hyperparameter for the \method. Shared hyperparameters are detailed below:

\begin{center}
\begin{tabular}{ll}
    \toprule
    Param & Value\\
    \midrule
    Batch Size & 128\\
    \# Training Batches & 50k \\
    Learning Rate & 5e-4\\
    Optimizer & AdamW\\
    Weight Decay & 0.1 \\
    Discount $\gamma$ & 0.99 \\
    Q Update Polyak Averaging Coefficient & 0.005\\
    Policy and Q Update Period & 1/train iter\\
    Nonlinearity & LeakyReLU(0.2) \\
    IQL Advantage Clipping & [0, 100]\\
    IQL Advantage Inverse Temperature $\beta$ & 5\\
    IQL Quantile $\tau$ & 0.8\\
    Action Chunking Length & 15\\
    \bottomrule
\end{tabular}
\end{center}

\paragraph{Fine-tuning details and hyperparameters.}
We collect 25 demonstrations for each downstream task and perform individual fine-tuning of the models for each task. In the case of the pre-trained models (SPRINT, L-BC composite, and L-BC primitive), we conduct 500 epochs of fine-tuning. As for the model without pre-training, we train 2000 epochs only on the downstream task demonstrations. The fine-tuning/training hyperparameters are identical to those for pre-training.

\subsection{Language-conditioned Behavior Cloning}
Our language-conditioned behavior cloning (L-BC) comparison method is inspired by and replicates BC-Zero~\citep{jang2021bc} and LangLfP~\citep{lynch2021language}. 
BC-Zero performs language imitation learning~\citep{perez_film_2017}, and both BC-Zero and LangLfP have an additional image/video-language alignment objective. In BC-Zero, their video alignment objective aligns language embeddings with videos of humans performing tasks related to those the BC-Zero robot agent trains on. LangLfP's image-language alignment objective allows their policy to accept both image and natural language goals as input due to only having a subset of their data labeled with hindsight language labels. As we don't have human videos of these tasks and our entire dataset is labeled with language labels, we do not add a video or image alignment objective.

Hyperparameters for the L-BC baseline are identical to the shared parameters above for both environments, where applicable.
\paragraph{ALFRED:} 
We implement L-BC by using the same architecture as described in the shared details section above with just a single transformer policy network that trains to maximize the log-likelihood of actions in the dataset. As our entire dataset consists of expert trajectories, this baseline ideally learns optimal actions for the instructions.

\paragraph{Real Robot:} L-BC is implemented with the action-chunked LSTM policy network to maximize log-likelihood of actions in the dataset as described in the real robot implementation details section above. Again the dataset consists of human expert trajectories so L-BC should learn optimal actions for the given instructions.

\subsection{Episodic Transformers}
Episodic Transformers (ET)~\citep{pashevich2021episodic} trains a transformer architecture on full sequences of ALFRED instructions with a behavior cloning objective. This is currently state of the art in the ``Seen Path-Length Weighted Success Rate'' evaluation metric on the ALFRED leaderboard.
We adopted the ET implementation from the official code repository.\\ \\
For fair comparison, we make a few modifications to make it as close as possible to \method\ and the baselines: 1) we train it on the same dataset as all baselines, so we do not generate new synthetic training data like the original implementation \citet{pashevich2021episodic} since it assumes access to an expert planner, 2) we encode visual frames with a Resnet-18 instead of Resnet-50 backbone, the same we use for all other models, 
3) we remove the high-level goal specification from the input text tokens as we do not assume access to those, and 
4) we train the model for longer to match the number of training steps for all methods.

\subsection{Actionable Models (AM)}
Actionable Models~\citep{actionablemodels2021arxiv} pre-trains a goal-conditioned Q function conditioned on randomly sampled image goals and also performs a goal-chaining procedure very similar to our skill chaining procedure. We implement AM by modifying the base IQL policy and critic networks to take in image goals instead of natural language embeddings as goals. These goals are provided in the same way as the observations as a sequence of 5 frames (the last 5 frames in the trajectory) processed by a frozen ResNet-18. 

To allow for fair comparison between our approach and AM, we implement AM with the same powerful offline RL algorithm, IQL~\citep{kostrikov2022offline}, used in our method. IQL ensures that the policy does not choose out-of-distribution actions by using advantage-weighted regression on in-distribution actions for policy extraction. With this, we found the conservative auxiliary loss AM adds to push down Q-values for out-of-distribution actions to be unnecessary and even hurtful to its overall performance, so we omit this additional loss term. 

\new{We also pre-train AM on the same long-horizon trajectories as those generated by \method\ during LLM-based skill aggregation. This ensures a fair comparison in terms of the types and lengths of tasks seen during pre-training.}

Finally, after consulting the authors of AM, we tried varying maximum trajectory lengths when sampling random goals. We found that allowing random goals to be sampled from anywhere within a trajectory resulted in the best zero-shot evaluation performance for AM, so our numbers are reported with this implementation detail.
\subsection{\method }
The implementation details of \method\ follow from the general discussions
at the top of this section. The key differences 
are in (1)~language model skill aggregation and (2)~cross-trajectory skill chaining, detailed below.

\paragraph{LLM Skill Aggregation.}
We perform LLM skill aggregation fully offline by iterating through every trajectory and aggregating sequences of adjacent primitive skill sub-trajectories. Assuming a trajectory with $N$ primitive skills, we select all ${N \choose 2}$ pairs of start and end skills and aggregate all instructions from start to end with the LLM. With 73k original language-annotated sub-trajectories in ALFRED, this procedure allows us to generate an additional 110k aggregated trajectories. We then add these trajectories to the original dataset and train on the entire set.

On our real-world robot dataset, we start with $\sim $6k language-annotated sub-trajectories and perform LLM skill aggregation on all pairs of trajectories directly next to each other (restricting to a maximum of 2 skills being aggregated at any time). We restrict aggregation in this manner because each trajectory contains many sub-trajectories of play-like data where many of the sub-trajectories are not related to each other. Aggregation doubles the size of our dataset to almost $\sim$13k trajectories.

\paragraph{Cross-trajectory skill chaining.}
We perform cross-trajectory skill chaining in-batch. Instead of sampling a second trajectory to perform chaining on, we simply permute the batch indicies to generate a set of randomly sampled second trajectories. Then, we perform a second loss function update, in addition to the original update on the sampled trajectories, with equal loss weighting, to apply the skill-chaining update. We apply the chaining procedures from Eq. \ref{eq:cross_chaining_reward} in-batch. Empirically, we found that cross-trajectory skill chaining works slightly better with the on-policy Value function obtained through IQL, therefore we use state values at the chaining targets instead of state-action Q-values.

\method-specific hyperparameters follow:

\begin{center}
\scalebox{1.0}{\begin{tabular}{ll}
    \toprule
    Param & Value\\
    \midrule
    LLM & LLAMA-13B~\citep{touvron2023llama}\\
    LLM Token Filtering Top-p & 0.9\\%\footnote{At each token generation step, \\only the highest probability tokens \\ with total probability mass that add up to the top-p are considered.} & 0.9\\
    LLM Token Sampling Temperature & 0.6\\
    \bottomrule
\end{tabular}}
\end{center}

\new{
\subsubsection{Cross-trajectory chaining preserves the MDP.}
\label{sec:appendix:cross-traj chaining discussion}
When performing cross-trajectory chaining using Eq.~\ref{eq:cross_chaining_reward}, special care must be taken to preserve the dynamics of the original Markov Decision Process (MDP). When chaining together two trajectories $\tau_A$ and $\tau_B$, we concatenate the two sentences of each trajectory together and relabel their rewards with Eq.~\ref{eq:cross_chaining_reward}.
The new language annotation used to chain together these trajectories is the concatenation of the two sentences, implying that the agent finishes skill (A) and then skill (B). However, we cannot concatenate the two trajectories together into one longer trajectory, as doing so would imply that the agent can instantaneously jump from the last state of skill (A) to the first state of skill (B), which may not be possible. Therefore, we instead treat the relabeled trajectories as separate trajectories with the same language annotation (lines 36 and 37 of Algorithm~\ref{alg:approach_summary}). \\ \\
However, this introduces two possible complications: 1) Language annotations differing in structure from those in the original dataset, and 2) Possible instruction ambiguity.
We detail how these complications are resolved in \method\ below: \\ \\
\begin{enumerate}
    \item \textbf{Language annotations differing in structure.} Language annotations produced by the chaining procedure will result in annotations that implicitly skip certain steps. For example, when chaining skill (A), ``make the bed,'' and skill (B), ``make a cup of coffee,'' the resulting chained annotation will be ``Make the bed. Make a cup of coffee.'' However to perform skill (B) the agent needs to first move to the kitchen from the bedroom to make the cup of coffee, which is skipped in this annotation. LLM-based skill aggregation (Section~ \ref{sec:llm_relabeling}) helps bridge this gap by summarizing long-horizon sequences while skipping certain implied steps. For example, one real LLM summary summarized the sequence: “\textit{1: Pick up the plaid pillow that is on the left end of the couch. 2: Place the pillow on the ottoman}” into the instruction “\textit{Place a plaid pillow on the ottoman,}” which skipped the step of picking up the pillow as it is implied that you must do so before placing the pillow down. Using the LLM augments our original dataset such that, in ALFRED, we have 2.5x the original data after performing offline skill aggregation, and in the real robot manipulation environment we have 2x the amount of original data. Therefore after performing LLM aggregation, there are many examples of similar instructions to those used for chained trajectories that imply certain steps without mentioning them explicitly.
    \item \textbf{Instruction ambiguity.} When chaining trajectories, there will be some ambiguity introduced as we do not have intermediate instructions for going from the last state of A to the initial state of B (obtaining these instructions requires additional human effort). This ambiguity is only present in the states of trajectory A, as when training on trajectory B, the agent can easily infer that the instructions for trajectory A are finished and the just follow the instructions relevant for trajectory B. We believe that the effects of the ambiguity on pre-training performance depends greatly on the given dataset. In complex and diverse environments, hindsight-labeled annotations should contain details specific to certain scenes, resolving this ambiguity. In ALFRED, the annotations usually contain information about the specific objects that the agent must interact with or locations that the agent must go to. For example, annotations for rinsing mugs typically are of the form “\textit{clean the MUG in the sink,}” or annotations for picking up a candle will often say something like “\textit{pick up the YELLOW CANDLE on the COUNTER,}” highlighting specific details regarding what the agent is supposed to do to complete the trajectory. 
\end{enumerate}
}

\subsection{SayCan}
In ALFRED experiments, we evaluate the performance of SayCan~\citep{saycan2022arxiv}, a top-down LLM-planning approach that breaks down a high-level task into
a sequence of steps that a language-conditioned policy can execute. SayCan does not perform any fine-tuning as it is not a pre-training method, instead we implement it by prompting a large language model to produce a probability distribution over the set of primitive skill instructions relevant for each task. Therefore it receives some privileged information over all of the other compared methods, including \method, about which primitive skills to choose from in each evaluation task.

Specifically, we use LLaMA-13B directly at test time to produce plans, the same model that we used to perform LLM skill relabeling for \method. The pre-trained policies are pre-trained on the same data as L-BC except that we also pre-train a value function to use with SayCan as it weighs skill predictions using both a pre-trained language-conditioned value function and the LLM-produced probabilities.

The prompt for SayCan, inspired by the prompt recommended in the original paper, and with the same number of examples as the one for \method, follows below:
\begin{mdframed}
\small
Robot: Hi there, I'm a robot operating in a house.
Robot: You can ask me to do various tasks and I'll tell you the sequence of actions I would do to accomplish your task.

Human: How would you put the box with keys on the sofa?\\
Robot:
1. Pick up the keys on the center table.
2. Put the keys in the box.
3. Pick up the box with keys.
4. Put the box with keys on the sofa close to the newspaper.\\

Human: How would you put a cooled slice of lettuce on the counter?\\
Robot:
1. Pick up the knife from in front of the tomato.
2. Cut the lettuce on the counter.
3. Set the knife down on the counter in front of the toaster.
4. Pick up a slice of the lettuce from the counter.
5. Put the lettuce slice in the refrigerator. take the lettuce slice out of the refrigerator.
6. Set the lettuce slice on the counter in front of the toaster.\\

Human: How would you put a book on the couch?\\
Robot:
1. Pick up the book on the table, in front of the chair.
2. Place the book on the left cushion of the couch.\\

Human: How would you put the cleaned fork in a drawer?\\
Robot:
1. Pick up the fork from the table.
2. Put the fork in the sink and fill the sink with water, then empty the water from the sink and remove the fork.
3. Put the fork in the drawer.\\

Human: How would you put the box of tissues on the barred rack?\\
Robot:
1. Take the box of tissues from the makeup vanity.
2. Put the tissues on the barred rack.
3. Take the box of tissues from the top of the toilet.
4. Put the tissues on the barred rack.\\

Human: How would you put a heated glass on the wooden rack?\\
Robot:
1. Pick up the glass from the sink.
2. Heat the glass in the microwave.
3. Put the glass on the wooden rack.\\

Human: How would you look at the box under the lamp light?\\
Robot:
1. Pick up the box from the far side of the bed.
2. Hold the box and turn on the lamp.\\

Human: How would you [HIGH LEVEL TASK DESCRIPTION]?\\
Robot:
1. [SKILL 1 EXECUTED SO FAR]
2. [SKILL 2 EXECUTED SO FAR]
...
N. \_\_\_\_
\end{mdframed}

\section{Dataset, Environment, and Task Details}
\label{sec:appendix:alfred}
\subsection{ALFRED}
\begin{figure*}[ht]
    \begin{subfigure}[t]{0.32\textwidth}
        \centering
        \includegraphics[width=\textwidth]{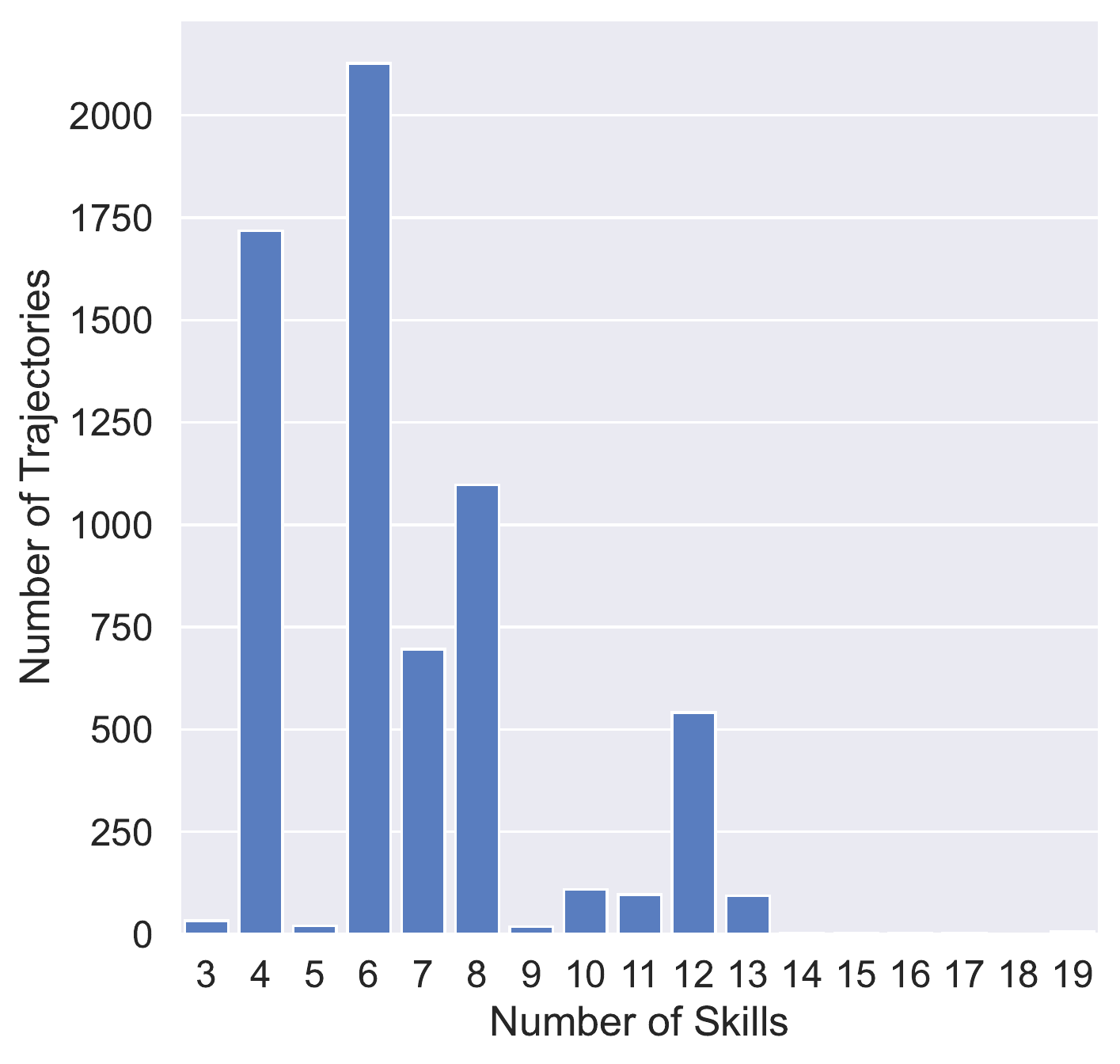}
        \caption{Skills per trajectory in the original ALFRED dataset.}
    \end{subfigure}
    ~
    \begin{subfigure}[t]{0.32\textwidth}
        \centering
        \includegraphics[width=\textwidth]{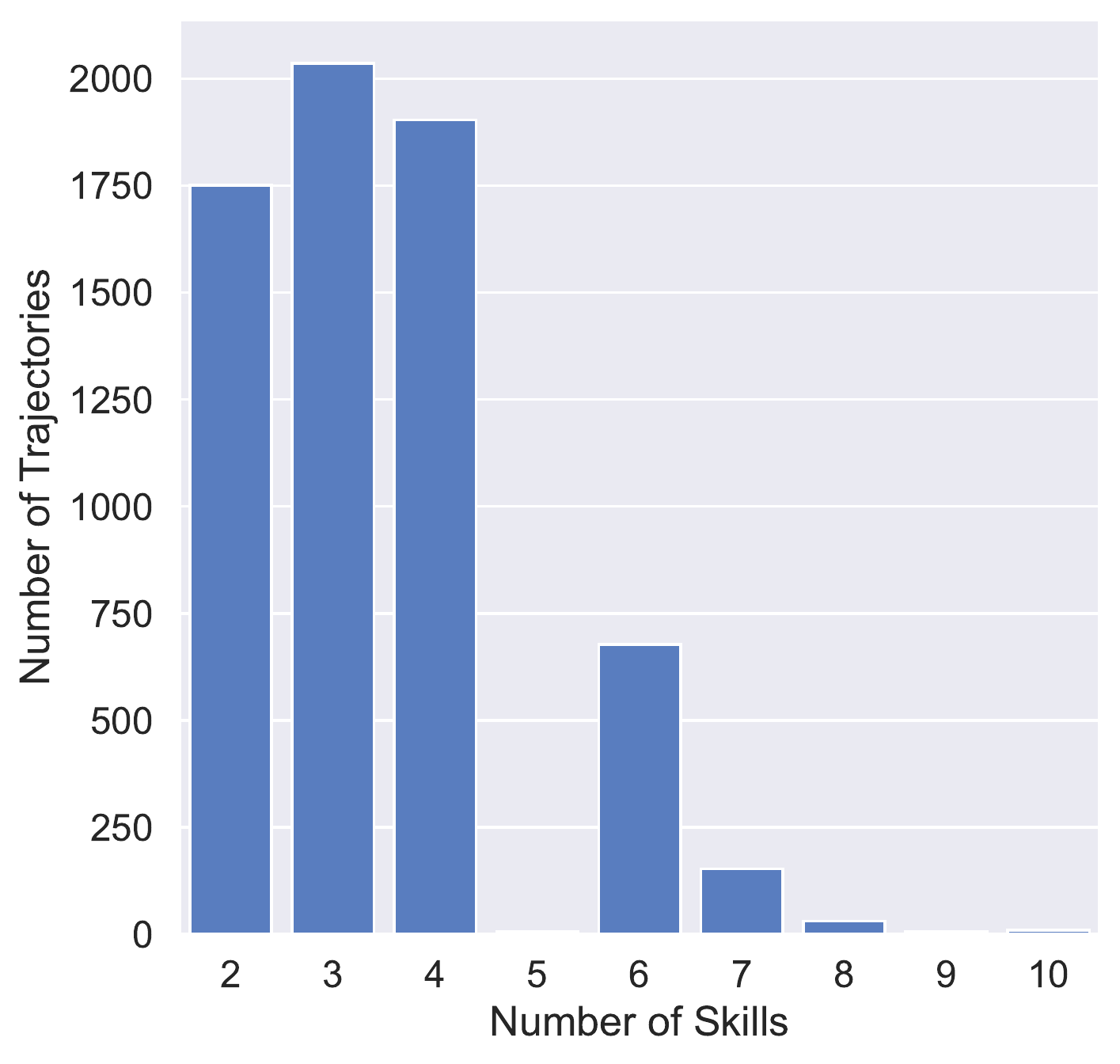}
        \caption{Skills per trajectory in the merged dataset.}
    \end{subfigure}
    ~
    \begin{subfigure}[t]{0.32\textwidth}
        \centering
        \includegraphics[width=\textwidth]{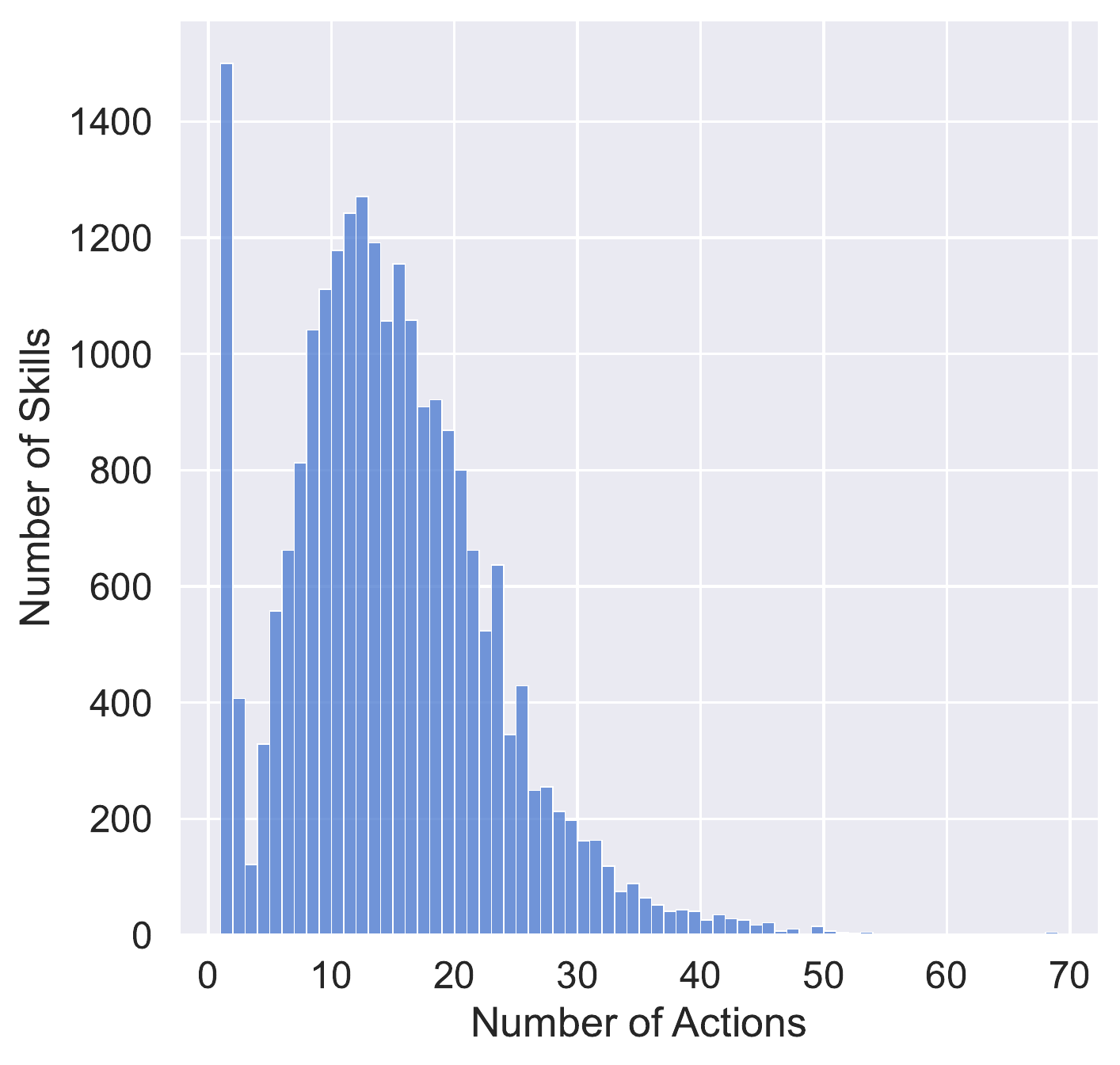}
        \caption{Actions per skill in the merged dataset.}
    \end{subfigure}
    \caption{\textbf{Left}: distribution of the number of skills in each trajectory in the original ALFRED dataset. \textbf{Middle}: distribution of skills per trajectory in the ``merged'' dataset with merged navigation skills. \textbf{Right}: distribution of number of actions per skill in the ``merged'' dataset.}
    \label{fig:skill_dist_in_traj}
\end{figure*}

\subsubsection{Dataset Details}
For training and evaluation we leverage the ALFRED benchmark and dataset~\citep{ALFRED20}. The ALFRED training dataset contains $\sim$6.6k trajectories collected by an optimal planner following a set of 7 high-level tasks with randomly sampled objects (\eg pick up an object and heat it). Each trajectory has at least three crowd-sourced sets of language instruction annotations. Each trajectory consists of a sequence of 3-19 individually annotated skills (see Figure~\ref{fig:skill_dist_in_traj}, left). This results in a total of 141k language-annotated skill trajectories. 

However, nearly half of the language instructions in the ALFRED dataset are navigation skill instructions like ``turn left, then look up and walk to the counter on the right''. To get a more balanced skill annotation dataset, we merge all navigation skills with the skill that immediately follows them, using only the annotation of the next skill. After this processing step, the resulting dataset contains 73k language-annotated primitive skill trajectories. After we merge the navigation skills, the average number of skills in each trajectory is 3.5 skills per trajectory (Figure~\ref{fig:skill_dist_in_traj}, middle), and the average number of actions in each skill is 14.3 (Figure~\ref{fig:skill_dist_in_traj}, right). %

\subsubsection{Evaluation Tasks}
\label{sec:appendix:eval_dataset}
\begin{table*}[ht]
    \centering
    \caption{Evaluation Task Specifics. Note that the ``number of env actions per task'' corresponds to the number of environment actions the ALFRED expert planner required to complete that task.}
    \label{tab:eval_task_specifics}
    \begin{tabular}{c|ccc}
    \toprule
        & \textit{EVAL\textsubscript{INSTRUCT}} & \textit{EVAL\textsubscript{LENGTH}} & \textit{EVAL\textsubscript{SCENE}}\\
        \midrule
    Number of Tasks & 100 & 20 & 10\\\midrule
    Task Lengths (\# primitive skills) & [1, 2, 3, 4, 5, 6, 7] & [7, 8] & [1, 2, 3, 4, 5]\\\midrule
    Min Number of Env Actions per Task & 1 & 34 & 2\\\midrule 
    Avg Number of Env Actions per Task & 39.1 & 60.9 & 46.6 \\\midrule
    Max Number of Env Actions per Task & 113 & 104 & 124\\\bottomrule
    \end{tabular}
\end{table*}

\begin{figure*}
    \centering
    \includegraphics[width=0.85\textwidth]{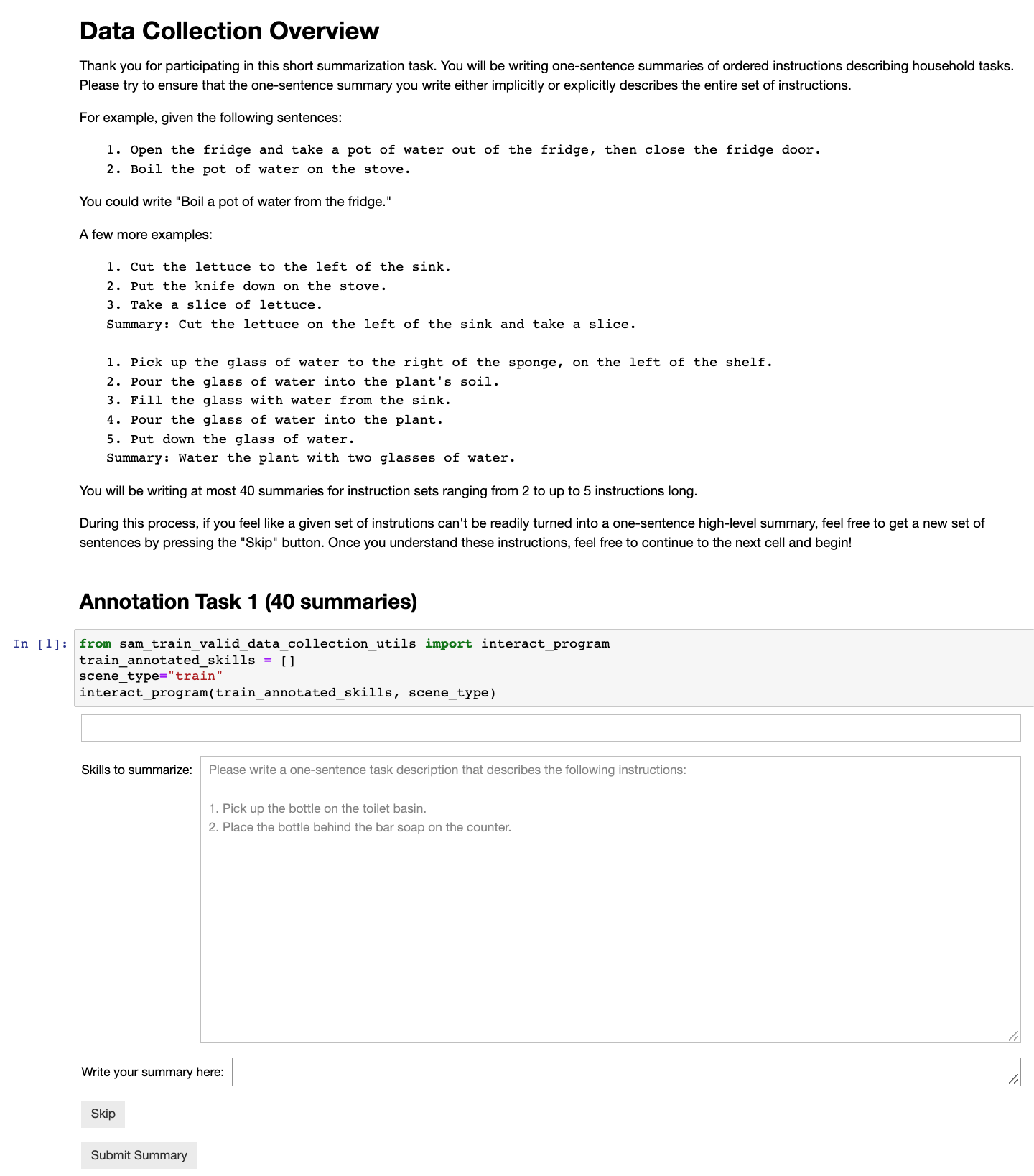}
    \caption{Data collection jupyter notebook page. \new{Note that there is a ``Skip'' button so that human annotators can skip an instruction sequence if they do not feel it is semantically meaningful or easy to summarize.}}
    \label{fig:data collection}
\end{figure*}
\paragraph{Overview.} We evaluate agents through zero-shot policy evaluation and finetuning on three sets of evaluation tasks in the ALFRED environment: (1) \textit{EVAL\textsubscript{INSTRUCT}} to measure the ability of pre-trained agents to execute semantically meaningful instructions at varied levels of abstraction, (2) \textit{EVAL\textsubscript{LENGTH}} to measure the ability of agents to chain behaviors across multiple trajectories to solve long tasks, and (3) \textit{EVAL\textsubscript{SCENE}} to evaluate generalization performance when finetuning to unseen household floor plans. We did not use the official ALFRED benchmark test sets to construct \textit{EVAL\textsubscript{SCENE}} since we require a task demonstration to compute how many subtasks the agent solved; these demonstrations are not given for the test set tasks. However, the tasks we evaluate on generally are designed to be representative of the tasks in the ALFRED test set: they test the agent on unseen instruction-scene combinations and consist of varied-length, compositional tasks. Like the ALFRED test set, our evaluation consists of long-horizon tasks that require sequential execution of multiple subtasks.

\paragraph{Collecting evaluation task data.} The ALFRED dataset provides high-level language annotations for each of the trajectories in the dataset. We could use these annotations as unseen task-instructions to evaluate our agents. However, we found that the different skills are not equally distributed across trajectories of different skill lengths, \eg most 2-skill trajectories perform pick-and-place tasks while tasks involving heating skills only appear in length 7+ trajectories. To allow evaluation with a less biased skill distribution, we create the \textit{EVAL\textsubscript{INSTRUCT}} task set by randomly choosing a trajectory from the ALFRED dataset \emph{and then randomly sampling a subsequence of skills \new{of a certain length}} from this trajectory. To obtain a high-level language instruction that summarizes this new subsequence, we crowd-source labels from human annotators. 
For labeling, each annotator is presented with a remotely hosted Jupyter notebook interface (see \myfig{fig:data collection}). %
Whenever we by chance sample a full ALFRED trajectory for annotation, we directly used the existing high-level annotation from the ALFRED dataset. %
We annotate 80 trajectories with human annotators and combine them with 20 randomly sampled single-skill trajectories, resulting in a total of 100 evaluation tasks (see Figure~\ref{fig:eval_100_example} for example instructions). \new{This results in 20 tasks of length 1 skills, 20 tasks of length 2 skills, 20 tasks of length 3 skills, 20 tasks of length 4 skills, and 20 tasks of lengths 5+ (5-7) skills.}

For \textit{EVAL\textsubscript{LENGTH}}, we randomly sampled 20 full trajectories from the ALFRED dataset that had sequences of 7 or 8 skills (10 of length 7, 10 of length 8) and removed these trajectories from the training dataset before performing LLM-based skill aggregation. This ensures AM and \method\ must perform 
skill chaining to solve these tasks by ensuring that there were valid sequences of skills to chain together to be able to solve these removed tasks. For example, assume a (shortened for clarity) sampled skill sequence is ``pick up apple,'' then ``put apple in microwave'', then ``slice the apple.'' Then, either Actionable Models or \method\ can chain together the sub-trajectory associated with ``pick up apple'' then ``put apple in microwave'' with the ``slice the apple'' sub-trajectory to solve this task. These trajectories all had annotations from ALFRED annotators, so we used those annotations directly (see Figure~\ref{fig:eval_chain_example} for example instructions).

Finally, for \textit{EVAL\textsubscript{SCENE}}, we collected a set of 10 full-length trajectories from the ALFRED ``valid-unseen'' dataset consisting of validation tasks in unseen floor plans. We collected 2 of each length from 1 through 5 for a total of 10 tasks by sampling random full-length trajectories from this dataset, with the exception of length 1 tasks (we just sample random skills to create length 1 tasks). As these are full trajectories, they already have human annotations from ALFRED, which we directly use as the task description (see Figure~\ref{fig:eval_unseen_example} for example instructions).

We list additional details about the tasks in each evaluation set in \mytable{tab:eval_task_specifics}.

Finally, we display 5 randomly sampled tasks, along with their human annotations, from each of our task sets in Figures \ref{fig:eval_100_example}, \ref{fig:eval_unseen_example}, and \ref{fig:eval_chain_example}. 
\begin{figure}[h]
\small
    \centering
    \begin{mdframed}
Skills to Summarize:
1: Grab the knife on the counter.
2: Place the knife in the sink then turn the faucet on so water fills the sink. Turn the faucet off and pick up the knife again.
3: Place the knife on the table to the left of the wooden bowl.\\
Annotator Summary: Wash the knife from the counter, put in on the table.\\

Skills to Summarize:
1: Pick up the blue book closest to your and the phone from the bed.
2: Turn on the lamp to take a look at the book in the light.\\
Annotator Summary: Examine the book by the light of a lamp.\\

Skills to Summarize:
1: Pick up yellow candle on counter.
2: Open cabinet, put candle in cabinet, close cabinet
3: Pick up yellow candle from toilet.\\
Annotator Summary: Move the candle from the sink to the cabinet under the sink, close it and and then pick the candle from the top of the toilet in front of you.\\

Skills to Summarize:
1: Pick the pot on the left side up from the stove.
2: Set the bowl and knife on the table next to the tomato.\\
Annotator Summary: Put the bowl with the knife in it next to the tomato.\\

Skills to Summarize:
1: Pick up the pen that's in front of you that's under the mug.
2: Put the pencil in the mug that was above it.
3: Pick up the mug with the pencil in it.\\
Annotator Summary: Put the pen into the mug and pick up the mug.\\
    \end{mdframed}
    \caption{Randomly sampled, human language instruction annotations from the \textit{EVAL\textsubscript{INSTRUCT}} task set.}
    \label{fig:eval_100_example}
\end{figure}
\begin{figure}[h]
\small
    \centering
    \begin{mdframed}
Skills to Summarize:
1: Pick up the lettuce on the counter.
2: Chill the lettuce in the fridge.
3: Put the chilled lettuce on the counter, in front of the bread.\\
Annotator Summary: Put chilled lettuce on the counter.\\

Skills to Summarize:
1: Pick up an egg from off of the kitchen counter.
2: Open the fridge, put the egg in to chill for a few seconds and then take it back out.
3: Place the cold egg in the sink.\\
Annotator Summary: Chill an egg and put it in the sink.\\

Skills to Summarize:
1: Pick up the butter knife off of the right side of the kitchen island.
2: Put the knife handle down in the frying pan that is on the front left burner of the stove.
3: Pick up the frying pan with the knife in it off of the stove.
4: Put the frying pan with the knife in it into the sink basin to the right of the potato.\\
Annotator Summary: Put a frying pan with a knife in it into the sink.\\

Skills to Summarize:
1: Take the pencil from the desk.
2: Put the pencil on the desk.\\
Annotator Summary: Take the pencil from the desk, put it on the other side of the desk.\\

Skills to Summarize:
1: Pick up the left pillow on the chair.
2: Put the pillow on the sofa right of the newspaper.
3: Pick up the pillow on the chair.
4: Put the pillow on the sofa left of the newspaper.\\
Annotator Summary: Place two pillows on a sofa.\\
    \end{mdframed}
    \caption{Randomly sampled, human language instruction annotations from the \textit{EVAL\textsubscript{SCENE}} task set.}
    \label{fig:eval_unseen_example}
\end{figure}

\begin{figure}[h]
\small
    \centering
    \begin{mdframed}
Skills to Summarize:
1: Pick up the knife in front of the lettuce.
2: Slice the apple in the sink with the knife.
3: Place the knife into the sink.
4: Pick up the sliced apple from the sink.
5: Place the apple slice into the pot on the stove.
6: Pick up the pot from the stove.
7: Pick ump the pot from the stove.
\\
Annotator Summary: Slice an apple for the pot on the stove and put the pot on the counter to the right of the door.\\

Skills to Summarize:
1: Take the apple from the counter in front of you.
2: Place the apple in the sink in front of you.
3: Take the knife by the sink in front of you.
4: Cut the apple in the sink in front of you.
5: Place the knife in the sink in front of you.
6: Take an apple slice from the sink in front of you.
7: Heat the apple in the microwave, take it out and close the microwave.
8: Place the apple slice in the sink in front of you.
\\
Annotator Summary: Place a warm apple slice in the sink.\\

Skills to Summarize:
1: Pick up the loaf of bread.
2: Put the bread on the counter above the spatula.
3: Pick up the knife that's above and to the right of the loaf of bread.
4: Cut the top half of the loaf of bread into slices.
5: Put the knife on the edge of the counter in front of you horizontally.
6: Pick up a slice of bread from the middle of the loaf.
7: Cook the bread in the microwave then take it out and close the microwave door.
8: Throw the cooked slice of bread away.
\\
Annotator Summary: Put a microwaved slice of bread in the oven.\\

Skills to Summarize:
1: Pick the knife up from off of the table.
2: Open the microwave, slice the potato, and close the microwave.
3: Open the microwave, place the knife inside of it, and close the microwave.
4: Open the microwave, pick up the potato slice inside, close the microwave.
5: Place the potato slice in the pan on the stove.
6: Pick up the pan from the stove.
7: Open the refrigerator, place the pan inside, and close the refrigerator.
\\
Annotator Summary: Move the pan from the stove top to inside the black refrigerator.\\

Skills to Summarize:
1: Pick up the red tomato on the counter to the right of the stove.
2: Put the tomato onto the island below the butter knife.
3: Pick up the butter knife off of the kitchen island.
4: Slice up the tomato on the kitchen island.
5: Place the butter knife onto the island to the right of the sliced tomato.
6: Pick up a tomato slice off of the kitchen island.
7: Open the fridge and put the tomato slice on the bottom shelf, then close the door, after a couple seconds open the fridge and remove the tomato slice then close the door.
8: Open the microwave door and place the tomato slice inside the microwave in front of the egg.
\\
Annotator Summary: Put a chilled tomato slice into the microwave.\\
    \end{mdframed}
    \caption{Randomly sampled, human language instruction annotations from the \textit{EVAL\textsubscript{LENGTH}} task set.}
    \label{fig:eval_chain_example}
\end{figure}

\paragraph{Online finetuning environment setup.}
During online-finetuning we initialize the agent in the same house floor plan as the trajectory the task was extracted from to ensure executability. During finetuning, we give each episode a time horizon of 2x the number of environment actions needed by the expert planner to solve the task. We give sparse sub-task rewards for each skill solved by the agent during the episode. Therefore for length 1 tasks, the agent can only be rewarded once before the episode ends, while for length 5 tasks, the episode terminates on the fifth reward signal. 

\subsection{Real Robot}
\label{sec:appendix:real_robot_env_setup}
Here we detail the dataset and evaluation tasks used for the real world tabletop environment experiments.
\paragraph{Dataset Details.}
Part of our data comes from data collected from prior work on the same arm setup ~\citep{dass2023jacoplay}, in addition to additional trajectories collected for this project.

In total, we collected 329 long-horizon trajectories, resulting in $\sim$6k individual ``primitive'' skills consisting of pick and place tasks such as \textit{``pick up the black bowl''} or \textit{``put the apple in the sink.''} These trajectories involve unique scene arrangements of different toy objects such as an apple, orange, black bowl, white plate, oven, sink, dish rack, etc. The total dataset size is 455,473 individual state-action pairs.

\paragraph{Evaluation Tasks.}
We formulate 3 unseen evaluation tasks requiring the completion of 2, 4, and 8 subtasks. These tasks are set in an environment configuration that has not been seen before in the training data, \ie the object combination is not present in the training data. For each of the tasks, we collect 25 demonstrations to finetune pre-trained policies for evaluation.

The three tasks are defined below:
\begin{enumerate}
    \item Bake bread in the oven (length 2): The robot must (1) pick up the bread, (2) place it in the oven.
    \item Serve heated milk in the bowl (length 4): The robot must (1) pick up the milk carton, (2) place it in the black bowl bowl, (3) pick up the bowl with the milk in it, (4) place the bowl in the oven.
    \item Serve milk in the bowl and butter and baked bread in the plate (length 8): The robot must: (1) pick up the milk carton, (2) put it in the black bowl, (3) pick up the butter stick, (4) put it in the plate, (5) pick up the bread, (6) bake the bread in the oven, (7) pick up the bread from the oven, (8) place the bread in the plate.
\end{enumerate}

\newpage
\section{Extended Experiments, Results, and Analysis}
\label{sec:appendix:experiments}
\begin{table*}[h]
\caption{\textit{EVAL\textsubscript{INSTRUCT}}  and \textit{EVAL\textsubscript{LENGTH}} eval dataset per-length and overall skill completion rates. See \mysec{sec:experiments} for experiment setup.}
\label{tab:zero_shot_numbers}
    \centering
\resizebox{\textwidth}{!}{%
    \begin{tabular}{lllllll}
    \toprule
& & AM & ET & L-BC & SayCan & \method\ \\\midrule 
 \multirow{8}{*}{\textit{EVAL\textsubscript{INSTRUCT}}} & Number of Completed Subtasks Overall & 0.82 $\pm$ 0.07 & 1.15 $\pm$ 0.14 & 0.39 $\pm$ 0.02 & 1.00 $\pm$ 0.12 & \textbf{1.94 $\pm$ 0.04} \\
& Length 1 Progress & 0.47 $\pm$ 0.06 & 0.75 $\pm$ 0.07 & 0.89 $\pm$ 0.07 & \textbf{0.94 $\pm$ 0.02} & 0.89 $\pm$ 0.04 \\
& Length 2 Progress & 0.75 $\pm$ 0.10 & 1.20 $\pm$ 0.18 & 0.66 $\pm$ 0.05 & 0.69 $\pm$ 0.22 & \textbf{1.52 $\pm$ 0.10} \\
& Length 3 Progress & 0.96 $\pm$ 0.28 & 1.61 $\pm$ 0.23 & 0.27 $\pm$ 0.08 & 0.61 $\pm$ 0.09 & \textbf{2.21 $\pm$ 0.03} \\
& Length 4 Progress & 0.56 $\pm$ 0.17 & 1.45 $\pm$ 0.25 & 0.05 $\pm$ 0.05 & 0.60 $\pm$ 0.18 & \textbf{2.36 $\pm$ 0.16} \\
& Length 5 Progress & 1.59 $\pm$ 0.53 & 0.76 $\pm$ 0.14 & 0.07 $\pm$ 0.05 & 0.57 $\pm$ 0.04 & \textbf{3.04 $\pm$ 0.24} \\
& Length 6 Progress & 1.20 $\pm$ 0.40 & 0.86 $\pm$ 0.94 & 0.05 $\pm$ 0.08 & 0.24 $\pm$ 0.08 & \textbf{2.87 $\pm$ 0.20} \\
& Length 7 Progress & 0.00 $\pm$ 0.00 & 0.00 $\pm$ 0.00 & 0.00 $\pm$ 0.00 & \textbf{0.40 $\pm$ 0.49} & 0.00 $\pm$ 0.00 \\
 \midrule
\multirow{3}{*}{\textit{EVAL\textsubscript{LENGTH}}} & Number of Completed Subtasks Overall & 1.71 $\pm$ 0.43 & 1.76 $\pm$ 0.14 & 0.07 $\pm$ 0.00 & 0.66 $\pm$ 0.08 & \textbf{4.40 $\pm$ 0.39}\\

& Length 7 Progress & 0.80 $\pm$ 0.16 & 0.78 $\pm$ 0.45 & 0.06 $\pm$ 0.05 & 0.50 $\pm$ 0.06 & \textbf{3.38 $\pm$ 0.43} \\
& Length 8 Progress & 2.62 $\pm$ 0.71 & 2.74 $\pm$ 0.55 & 0.26 $\pm$ 0.18 & 1.50 $\pm$ 0.24 & \textbf{5.25 $\pm$ 0.64 }\\
\bottomrule

\end{tabular}}

\end{table*}

Here, we present additional results complementary to the experiments in the main paper in \mysec{sec:experiments}. 
We present and analyze LLM annotation examples in \mysec{sec:appendix:experiments:summaries}.

\subsection{LLM Summary Examples}
\label{sec:appendix:experiments:summaries}

\begin{figure*}[h]
    \centering
    \begin{mdframed}
Skills to summarize:
1. Pick the clock up from the desk.
2. Turn the lamp on. \\
\textbf{OPT-350m Summary:} Look at the clock.\\
\textbf{OPT-1.3b Summary:} Look at the clock from the far side of the bed under the lamp light.\\
\textbf{LLaMA-13b Summary:} Look at the clock from the desk under the lamp light.\\ \\

Skills to summarize: 
1. Take the plate from the cabinet to the left of the stove. 
2. Warm the plate in the microwave, above the stove.
3. Put the plate in the cabinet.\\
\textbf{OPT-350m Summary:} Place the plate in the microwave.\\
\textbf{OPT-1.3b Summary:} Put the plate on the left side of the table.\\
\textbf{LLaMA-13b Summary:} Put a warm plate in the cabinet above the stove.\\ \\

Skills to summarize:
1. Open the microwave, take out the potato and shut the door.
2. Put the potato in the sink, turn the water on and off again. Take the potato out of the sink.
3. Place the potato on the counter, to the right of the sink.\\
\textbf{OPT-350m Summary:} Put the potato on the counter, to the left of the sink.\\
\textbf{OPT-1.3b Summary:} Open the microwave, take out the potato, and close the door.\\
\textbf{LLaMA-13b Summary:} Put a potato on the counter, next to the sink.\\ \\
    \end{mdframed}
    \caption{Comparison of OPT-350m, OPT-1.3b~\citep{OPT}, and LLaMA-13b summaries on 3 randomly sampled sequences from our dataset. In general, as the model size increases, the summary becomes better. LLaMA-13b summaries as a whole tend to be more descriptive and accurate. The smaller models tend to regurgitate information from the original skills incorrectly, such as OPT-350m summarizing the third example as putting a potato ``to the left of the sink'' when the original skill stated ``to the right of the sink.''}
    \label{fig:llm_comparison}
\end{figure*}

We randomly sample 3 LLAMA-13b task summaries produced while performing
skill aggregation in ALFRED (explained in \mysec{sec:llm_relabeling}) using the prompt in \myfig{fig:llm_prompt} and display them in Figure~\ref{fig:llm_comparison} along with summaries from OPT-350m and OPT-1.3B~\citep{OPT}, 350M and 1.3B parameter open-source models for comparison. After analyzing many more examples, we see that LLaMA-13b generally provides fitting high-level summaries for most sequences by skipping over implied sub-tasks (although it sometimes also skips over important sub-tasks, likely due to the prompt). The other smaller models, which are also pre-trained on smaller corpora, tend to produce worse summaries and make up details more often. %

\subsection{Qualitative Comparison Results}
\label{sec:appendix:video_demo}
Here we display and analyze some qualitative task execution examples from ALFRED and our real robot environment.
\subsubsection{ALFRED}
\paragraph{Zero-shot evaluation.} 
We compare \method, AM, and L-BC zero-shot evaluation results on long \textit{EVAL\textsubscript{LENGTH}} tasks in \myfig{fig:zero_shot_qualitative_comparisons}. In general, \method\ is able to make substantially more progress on \textit{EVAL\textsubscript{LENGTH}} tasks as it leverages the large language model to generate longer-horizon, semantically meaningful pre-training tasks and performs cross-trajectory chaining to learn to chain its existing dataset tasks. In the visualized examples, \method\ is able to understand and successfully execute many of the sub-tasks implied but not directly stated by the natural language task instruction. L-BC makes very little progress on these tasks, not even understanding what the first sub-task to complete should be as the task annotation is out of distribution from what it saw while training. Finally, AM is able to make some progress on some of these tasks due to its long-horizon goal pre-training objective. However, this is less effective than our language-conditioned pre-training in such zero-shot evaluations.

We show some example plans generated by SayCan, two that did not complete the task and one that did, in Figure~\ref{fig:saycan_plan}. While SayCan can generate correct plans for certain tasks, the plans generated are subject to failures by the LLM to pick the correct skill.

\begin{figure}
    \centering
    \begin{mdframed}[frametitle={Task: Cut a potato and put a slice in the sink. (Fail)}]
         \begin{subfigure}[t]{0.49\textwidth}
            \textsc{Ground Truth}
            \begin{enumerate}
                \item Pick up the knife from the sink. 
                \item Cut the potato on the second shelf in the fridge. 
                \item Put the knife back in the sink. 
                \item Take a potato slice from the sink. 
                \item Put the slice in the sink.
            \end{enumerate}
         \end{subfigure}
         \begin{subfigure}[t]{0.49\textwidth}
            \textsc{SayCan Generated Plan}
            \begin{enumerate}
                \item Pick up the knife from the sink. 
                \item Put the knife back in the sink. 
            \end{enumerate}
         \end{subfigure}
    \end{mdframed} 
    \begin{mdframed}[frametitle={Task: Move all computers onto the white dresser. (Fail)}]
         \begin{subfigure}[t]{0.49\textwidth}
            \textsc{Ground Truth}
            \begin{enumerate}
            \item Place the computer on the white dresser. 
            \item Close the computer and pick up the computer from the green bed.
            \item Place the computer on the white dresser.
 
            \end{enumerate}
         \end{subfigure}
         \begin{subfigure}[t]{0.49\textwidth}
            \textsc{SayCan Generated Plan}
            \begin{enumerate}
            \item Place the computer on the white dresser. 
            \item Place the computer on the white dresser. 
            \end{enumerate}
         \end{subfigure}
    \end{mdframed} 
    \begin{mdframed}[frametitle={Task: Move the phone from the dresser to the bed. (Success)}]
         \begin{subfigure}[t]{0.49\textwidth}
            \textsc{Ground Truth}
            \begin{enumerate}
            \item Take the blue cell phone off of the dresser. 
            \item Put the blue cell phone on the bed.
            \end{enumerate}
         \end{subfigure}
         \begin{subfigure}[t]{0.49\textwidth}
            \textsc{SayCan Generated Plan}
            \begin{enumerate}
            \item Take the blue cell phone off of the dresser. 
            \item Put the blue cell phone on the bed.
            \end{enumerate}
         \end{subfigure}
    \end{mdframed} 
    \caption{Example plans from SayCan~\citep{saycan2022arxiv} evaluated on \emph{EVAL\textsubscript{INSTRUCT}}. For longer tasks SayCan has a higher probability of generating any single incorrect step in a plan, leading to planning failures that will prevent the language-conditioned policy from completing the task.}
    \label{fig:saycan_plan}
\end{figure}

\paragraph{Finetuning.} We finetune \method, AM, and L-BC on \textit{EVAL\textsubscript{SCENE}} tasks, in household floorplans that were never seen while training, and visualize qualitative policy rollout examples \textit{after finetuning} in \myfig{fig:finetuning_qualitative_comparisons}. In general, \method\ is able to finetune to longer-horizon tasks while AM and L-BC both struggle with making progress on longer-horizon tasks despite receiving rewards for every completed sub-task. \method's ability to complete more sub-tasks on many of the longer-horizon tasks is demonstrated in Figure~\ref{fig:qualitative_comparisons4}, while a case in which both \method\ and AM make partial progress throughout finetuning is demonstrated in Figure~\ref{fig:qualitative_comparisons5}. We believe that AM has more trouble finetuning on these tasks than \method\ because the task specification for AM (goal images) is out of distribution; pre-training on \textit{language} tasks with \method\ allows agents to more easily learn longer-horizon behaviors as the task specifications may still be in-distribution of the pre-training tasks that LLM skill-aggregation and skill chaining produce.

\begin{figure*}
    \centering
    \includegraphics[width=\textwidth]{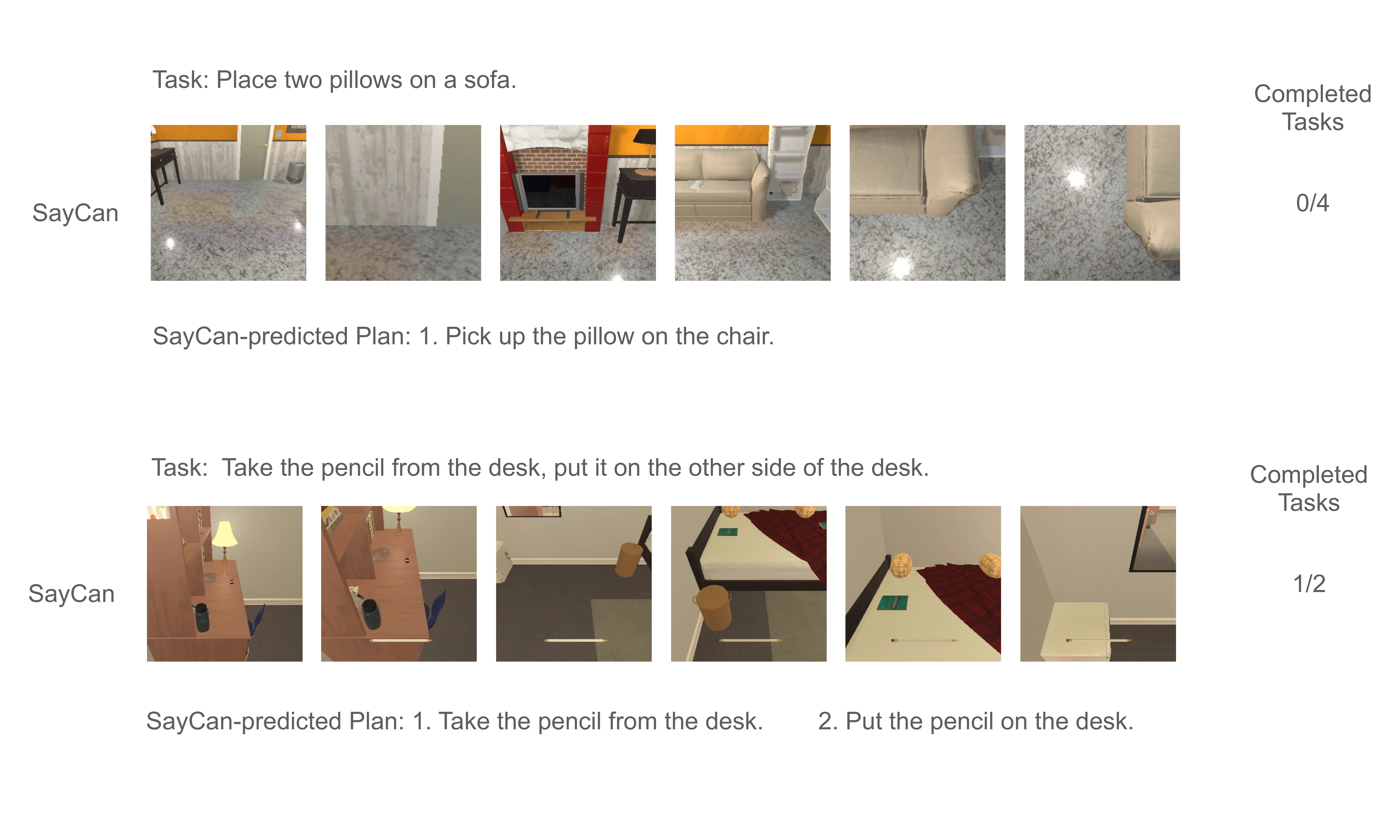}
    \caption{Rollouts of SayCan on \emph{EVAL\textsubscript{SCENE}}. In these examples, SayCan predicts correct plan steps until the policy suffers from execution errors as it is not fine-tuned for the unseen environments.}
    \label{fig:saycan_rollouts}
\end{figure*}

\begin{figure*}[p]
     \centering
     \begin{subfigure}[b]{0.9\textwidth}
         \centering
         \includegraphics[width=0.9\textwidth]{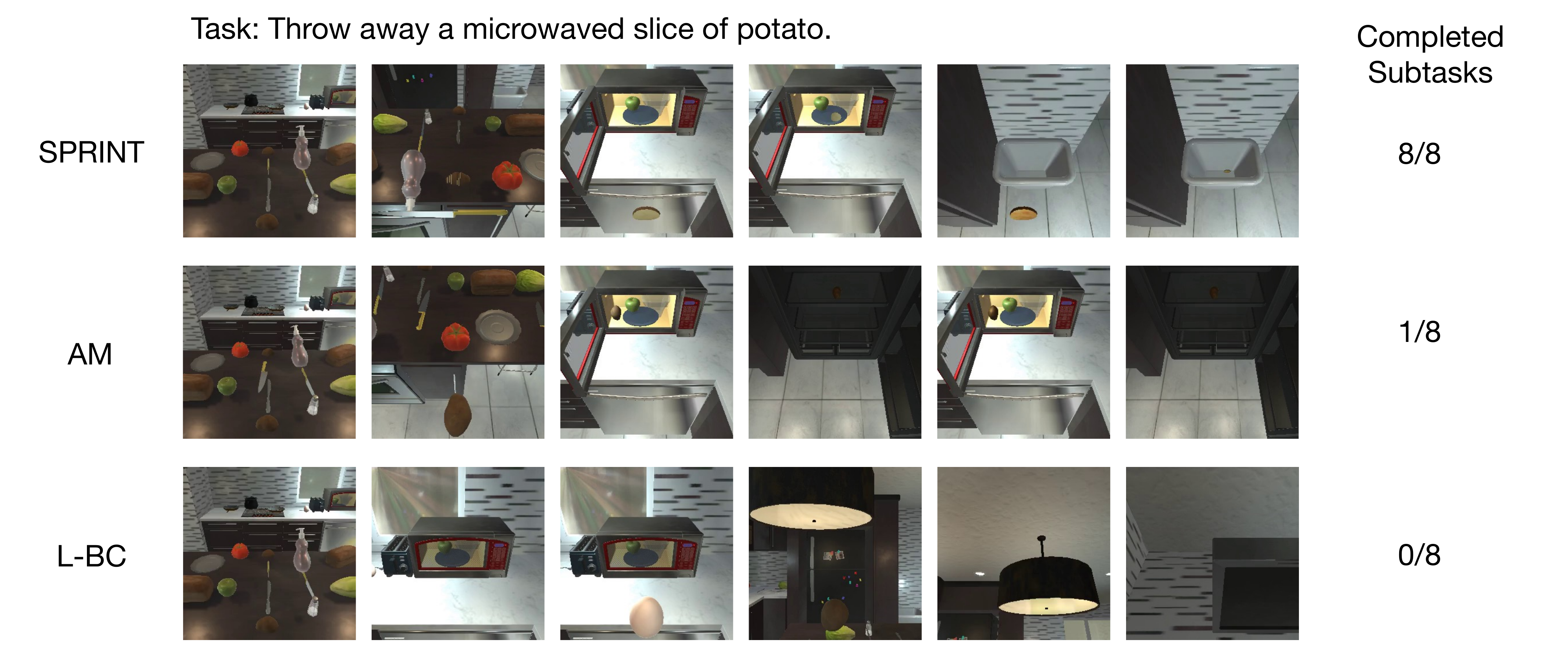}
         \caption{\method\ successfully solves this task, while AM fails to slice the potato and repetitively iterates between putting the potato in the fridge and microwave. L-BC fails even to pick up the potato, as the task annotation does not directly describe picking up a potato.}
         \label{fig:qualitative_comparisons1}
     \end{subfigure}
     \vfill
     \begin{subfigure}[b]{0.9\textwidth}
         \centering
         \includegraphics[width=0.9\textwidth]{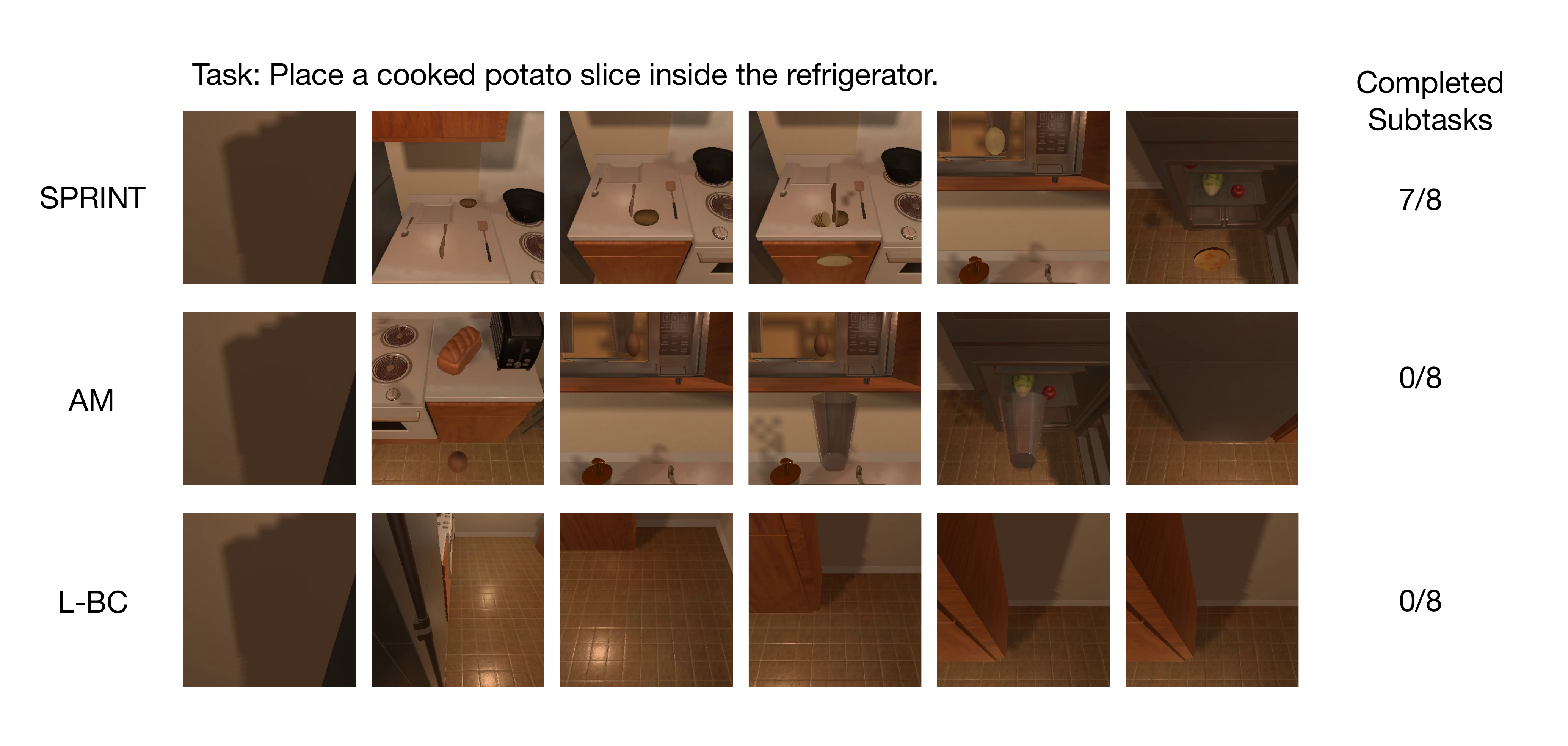}
         \caption{\method\ nearly solves this task, while AM picks up an egg instead of a potato. L-BC picks up random objects not related to the annotation.}
         \label{fig:qualitative_comparisons2}
     \end{subfigure}
     \vfill
     \begin{subfigure}[b]{0.9\textwidth}
         \centering
         \includegraphics[width=0.9\textwidth]{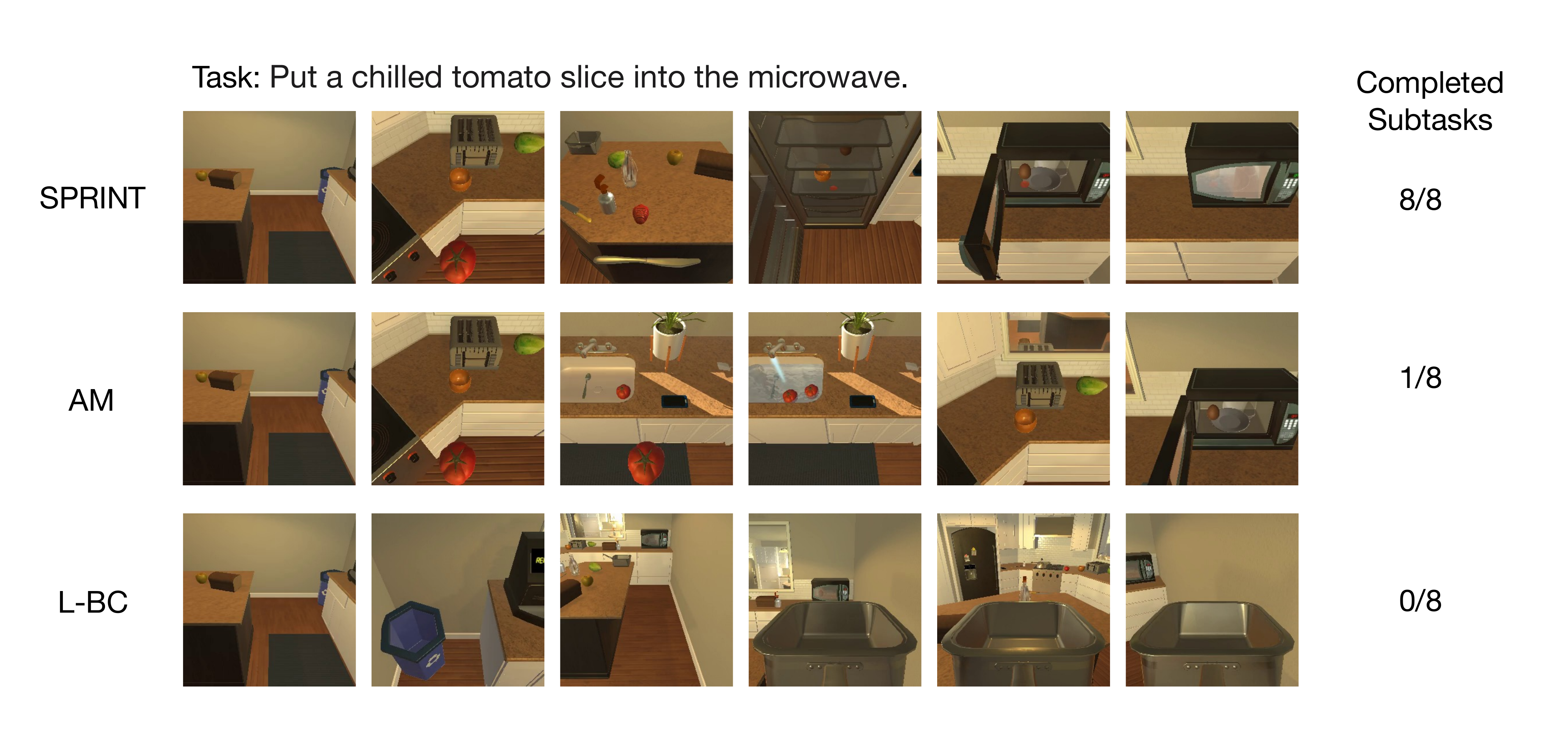}
         \caption{\method\ completes the entire task. AM picks up the tomato but fails to put it down onto the counter and slice it. L-BC aimlessly wanders and picks up random objects.}
         \label{fig:qualitative_comparisons3}
     \end{subfigure}
        \caption{Visualizations of zero-shot policy rollouts on three tasks in the \textit{EVAL\textsubscript{LENGTH}} task set.}
        \label{fig:zero_shot_qualitative_comparisons}
\end{figure*}

\begin{figure*}[p]
    \centering
    \begin{subfigure}[b]{0.9\textwidth}
    \includegraphics[width=0.9\textwidth]{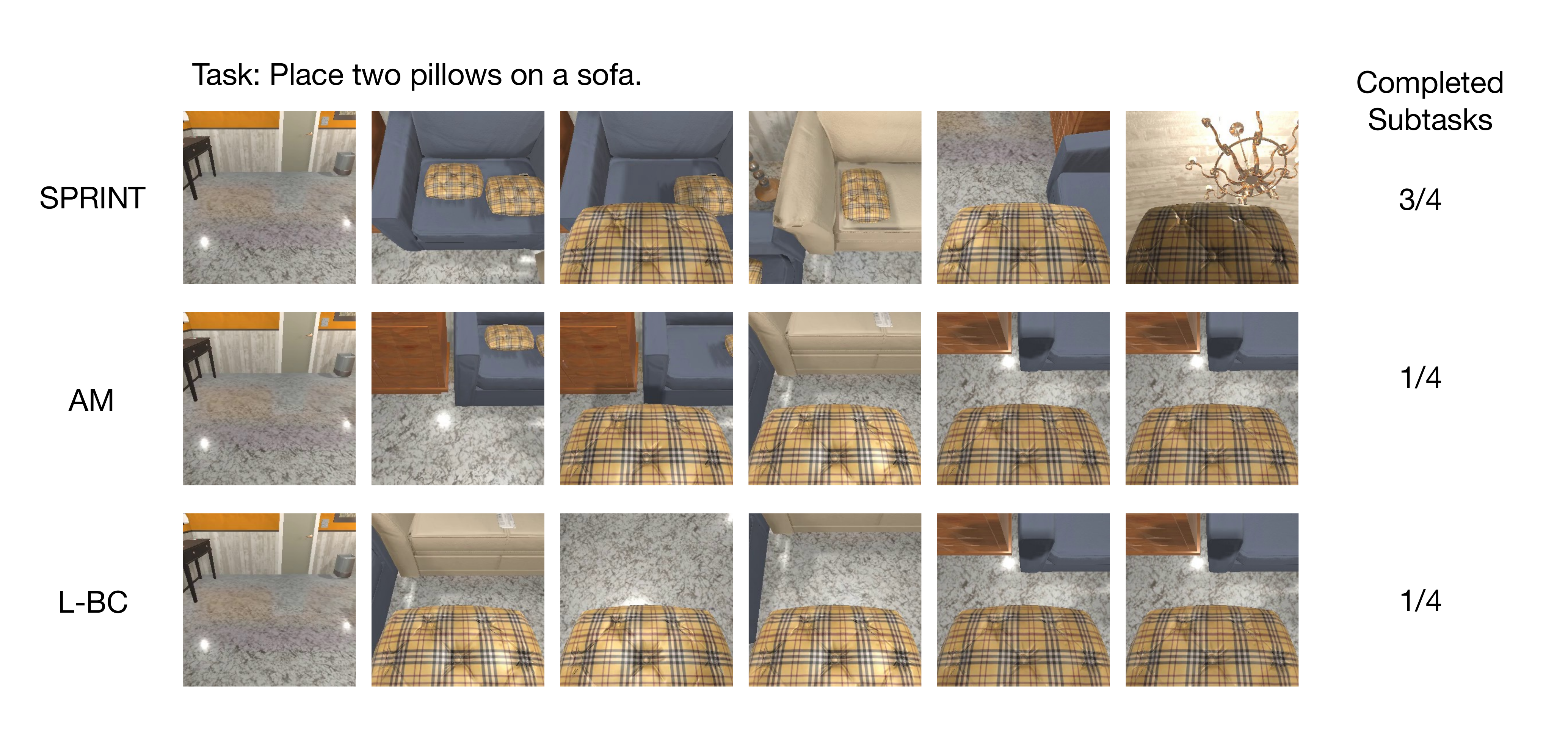}
    \caption{\method\ picks up and places one of the pillows on the sofa, and picks up the second but does not manage to place the second on the sofa, thus completing 3/4 subtasks. AM and L-BC both learn to pick up a pillow but never learned to place it in the correct spot.}
    \label{fig:qualitative_comparisons4}
    \end{subfigure}
    \begin{subfigure}[b]{0.9\textwidth}
    \includegraphics[width=0.9\textwidth]{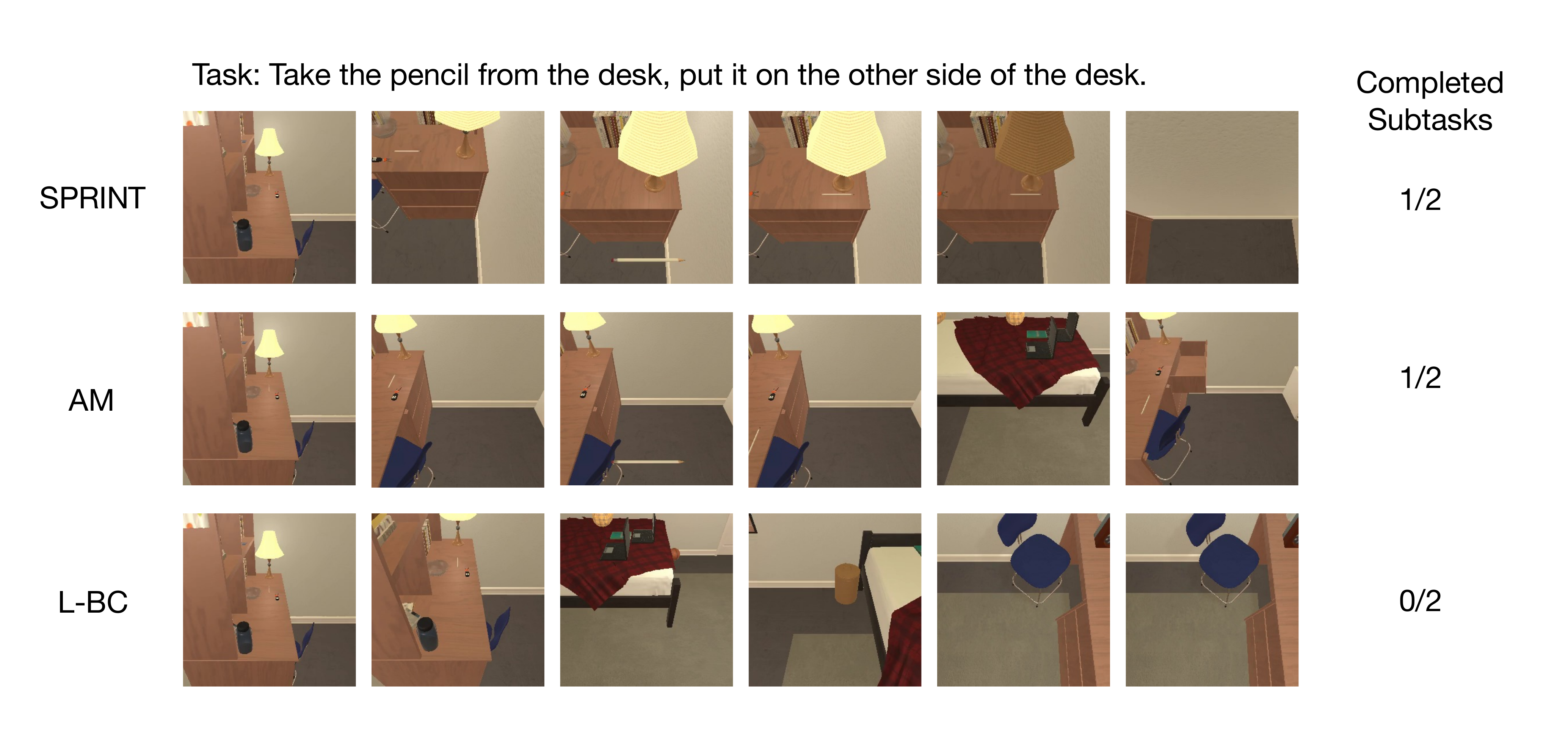}
    \caption{\method\ and AM both learn to pick up a pencil from the desk, although neither manage to put the pencil down in the correct place ``on the other side of the desk.'' Meanwhile, L-BC never picks up the pencil.}
    \label{fig:qualitative_comparisons5}
    \end{subfigure}
    \caption{Visualizations of policy rollouts on two tasks in the \textit{EVAL\textsubscript{SCENE}} task set, after finetuning each method. These floor plans were originally unseen to all agents until finetuning.}
    \label{fig:finetuning_qualitative_comparisons}
\end{figure*}

We do not fine-tune SayCan as it is not a pre-training/fine-tuning method. This makes it susceptible to both planning and policy execution failures in the unseen environments in \emph{EVAL\textsubscript{SCENE}}. We demonstrate policy execution failures in Figure~\ref{fig:saycan_rollouts}.

\subsubsection{Real Robot}
\begin{figure*}[p]
    \centering
    \includegraphics[width=1\textwidth]{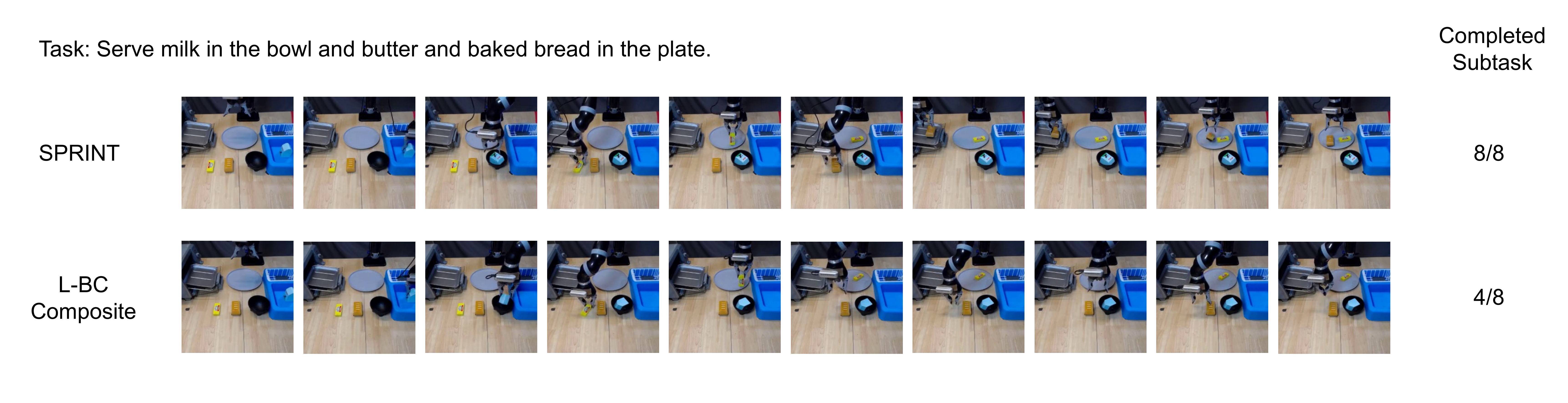}
    \caption{\method\ picks up the correct objects successfully and places in the right place accurately, with the same order shown in the demonstration. L-BC composite model does the right thing on the milk diary and butter diary but is not able to finish any skills with the long bread.}
\label{fig:real_robot_qualitative_comparisons}
\end{figure*}

We visualize evaluation rollouts after finetuning on our most difficult, length 8 task, \textit{``Serve milk in the bowl and butter and baked bread in the plate,''} in Figure~\ref{fig:real_robot_qualitative_comparisons}.
We display an example comparison between \method\ and L-BC composite, the best-performing L-BC baseline in which the fine-tuned SPRINT model successfully follows and accomplishes the skills in the demonstrated long-horizon sequences. 
The L-BC composite agent finishes the first four skills before encountering confusion about the subsequent skill: pick up the long bread. This comparison reveals that the L-BC composite model exhibits proficiency in completing some skills but overall does struggle with long-horizon tasks.
Empirically in our evaluations, we saw that this baseline exhibits greater variance than \method\ among its evaluation runs, sometimes only executing 2 skills and other times finishing 8.

\end{document}